\documentclass[journal,twoside,web]{ieeecolor2}
\usepackage{generic}
\usepackage{cite}
\usepackage{amsmath,amssymb,amsfonts}
\usepackage{algorithmic}
\usepackage{graphicx}
\usepackage{textcomp}
\usepackage{xcolor}
\usepackage{subcaption}
\usepackage{tabularx} 
\usepackage{booktabs} 

\usepackage{multirow}
\usepackage{multicol}

\usepackage{tikz}

\newcommand\submittedtext{%
  \footnotesize This work has been submitted to the IEEE for possible publication. Copyright may be transferred without notice, after which this version may no longer be accessible.}

\newcommand\submittednotice{%
\begin{tikzpicture}[remember picture,overlay]
\node[anchor=south,yshift=10pt] at (current page.south) {\fbox{\parbox{\dimexpr0.65\textwidth-\fboxsep-\fboxrule\relax}{\submittedtext}}};
\end{tikzpicture}%
}

\newcommand{\myparagraph}[1]{\smallskip\noindent\textbf{#1.}}
\newcommand{\mytitle}[1]{\smallskip\noindent\textbf{#1:}}

\newcommand{\revision}[1]{{\color{black}{#1}}}
\newcommand{\re}[1]{{\color{black}{#1}}}
\newcommand{\ree}[1]{{\color{black}{#1}}}

\def\onedot{.}
\def\eg{\emph{e.g}\onedot} 
\def\ie{\emph{i.e}\onedot}

\def\etal{\emph{et al}\onedot}

\def\method{{MINDS}}
\def\method{{CAMI-2DNet}}
\def\papertitle{{Computerized Assessment of Motor Imitation for Distinguishing Autism in Video}}

\begin{document}
\title{\papertitle\ (\method)}
\author{
Kaleab A. Kinfu, \IEEEmembership{Graduate Student Member, IEEE}, Carolina Pacheco,
Alice D. Sperry, Deana Crocetti, Bahar~Tun\c{c}gen\c{c}, Stewart H. Mostofsky, Ren\'e Vidal, \IEEEmembership{Fellow, IEEE}
\thanks{K. Kinfu and R. Vidal are with the Center for Innovation in Data Engineering and Science at the University of Pennsylvania. C. Pacheco is with the Department of Biomedical Engineering at Johns Hopkins University. A. Sperry, D. Crocetti and S. Mostofsky are with the Center for Neurodevelopmental and Imaging Research at the Kennedy Krieger Institute. S. Mostofsky is also affiliated with the Department of Neurology and the Department of Psychiatry and Behavioral Sciences at the Johns Hopkins University School of Medicine.
B. Tun\c{c}gen\c{c} is with the Department of Psychology at the Nottingham Trent University. This work was supported by NSF grants 2124276, 2124277, and 2430816, and by NIH grant R01NS135613.
}}

\maketitle

\submittednotice
\begin{abstract}
Motor imitation impairments are commonly reported in individuals with autism spectrum conditions (ASCs), suggesting that motor imitation could be used as a phenotype for addressing autism heterogeneity. Traditional methods for assessing motor imitation are subjective and labor-intensive, and require extensive human training. Modern Computerized Assessment of Motor Imitation (CAMI) methods, such as CAMI-3D for motion capture data and CAMI-2D for video data, are less subjective. However, they rely on labor-intensive data normalization and cleaning techniques, and human annotations for algorithm training. To address these challenges, we propose \method, a scalable and interpretable deep learning-based approach to motor imitation assessment in video data, which eliminates the need for ad hoc normalization, cleaning, and annotation. \method\ uses an encoder-decoder architecture to map a video to a motion representation that is disentangled from nuisance factors such as body shape and camera views. To learn a disentangled representation, we employ synthetic data generated by motion retargeting of virtual characters through the reshuffling of motion, body shape, and camera views, as well as real participant data. To automatically assess how well an individual imitates an actor, we compute a similarity score between their motion encoding, and use it to discriminate individuals with ASCs from neurotypical (NT) individuals. Our comparative analysis demonstrates that \method\ has a strong correlation with human scores while outperforming CAMI-2D in discriminating ASC vs NT children. Moreover, \method\ performs comparably to CAMI-3D while offering greater practicality by operating directly on video data and without the need for ad hoc normalization and human annotations.
\end{abstract}


\section{Introduction}
\label{sec:introduction}
\IEEEPARstart{A}{utism} spectrum conditions (ASCs) are defined by core symptoms of social-communicative difficulties,  restricted interests, and repetitive behaviors. However, the considerable heterogeneity in ASCs creates challenges for efficient diagnosis and treatment options~\cite{Masi2017AnOO}. Based on a growing amount of neuroscience and behavioral research, one promising way of addressing this heterogeneity is through precise quantification of motor imitation impairments, which are highly prevalent in autism~\cite{Bhat2020MotorII}. Motor imitation skills are fundamental for socialization, communication, and acquiring essential skills, particularly during early development and in social interactions~\cite{Over2013TheSS}. Motor imitation impairments seem specific to ASCs, and not shared with other highly co-occurring conditions, such as Attention-Deficit/Hyperactivity Disorder (ADHD)~\cite{Santra2025EvaluatingCA}, which are typically challenging to distinguish in clinical settings. Importantly, impairments in motor imitation are associated with core autism symptoms and brain mechanisms involved in social communication and learning~\cite{Lidstone2021MovingTU, Nebel2016IntrinsicVS}.

Human Observation Coding (HOC) has been the standard method for assessing motor imitation. HOC relies on trained human coders to directly observe and evaluate individuals' imitation skills~\cite{Edwards2014AMO}. Although it provides valuable insights into specific imitation challenges indicative of ASCs, it has several limitations. First, it is inherently subjective, since it depends~on\\ 
human judgment, potentially introducing biases and inconsistencies. In addition, HOC is labor-intensive, requiring significant time, effort, and trained personnel for accurate analysis and behavior coding. These limitations significantly hinder its scalability and practicality in clinical and home settings.

With the increasing prevalence of ASCs~\cite{Chiarotti2020EpidemiologyOA, Lyall2017TheCE} and the growing demand for early and accurate assessments, there is a need for automated and objective assessment tools that address the challenges associated with HOC. The development of tools that are effective and widely applicable offers several potential advantages, including efficiency, objectivity, and scalability. However, the development of such tools faces several challenges, including (i) the diverse range and complex nature of human actions involved in imitation, (ii) the trade-off between sensitivity and specificity in recognizing atypical imitation patterns, and (iii) the need to ensure the adaptability of the tool across diverse settings. 

Several automated methods have been proposed to address these challenges~\cite{Michelet2012AutomaticIA, Schmidt2012MeasuringTD, Paxton2012FramedifferencingMF, Chaudhry2009HistogramsOO, Ravichandran2009ViewinvariantDT}. 
Among these, motion-capture-based methods, which rely on precise 3-dimensional (3D) motion data, have been proven to be effective.
For instance, Computerized Assessment of Motor Imitation (CAMI-3D)~\cite{Tungen2021ComputerisedAO} uses a Kinect Xbox cameras to collect 3D motion data and applies Dynamic Time Warping (DTW)~\cite{Bellman1959OnAC} and linear regression to produce a similarity score that takes into account variations in both motion trajectories and timing discrepancies.

CAMI-3D has shown promising results in autism, demonstrating high test-retest reliability and surpassing the performance of HOC in effectively discriminating children with ASCs from neurotypical (NT) children as well as from children with ADHD, a highly prevalent condition that is both a differential diagnosis of ASC and a frequent co-occurring diagnosis~\cite{Santra2025EvaluatingCA}. However, despite its promising results, CAMI-3D has several limitations. First, it relies on specialized hardware, such as Kinect or similar 3D cameras, which may restrict its scalability and applicability in settings like homes or clinics. Second, obtaining accurate motion coordinates requires extensive manual frame-by-frame data cleaning to obtain accurate motion coordinates.
CAMI-3D also depends on hand-crafted normalization techniques to handle variations in body shape (\eg, height, limb length) and slight differences in camera angles. While these techniques may be effective in controlled studies, they may not generalize well to diverse anatomical and pose variations encountered in real-world scenarios. Third, HOC annotations are needed for training, thus requiring continued human input for new action sequences.

To address the dependence on costly 3D motion capture devices and enhance scalability, advances in 2D pose estimation techniques~\cite{Cao2018OpenPoseRM, Sun2019DeepHR, Kinfu2023EfficientVT} can be utilized. These methods accurately detect and track skeletal joints from video data captured by widely accessible 2D cameras. Motivated by these advances, the CAMI-2D method~\cite{Lidstone2021AutomatedAS} employs an off-the-shelf pose estimation network, OpenPose~\cite{Cao2018OpenPoseRM}, to extract 2D joint trajectories from videos. Similar to CAMI-3D, CAMI-2D compares these joint trajectories using DTW and computes imitation scores through linear regression. However, CAMI-2D inherits CAMI-3D limitations, such as the need for hand-crafted normalization and ongoing human annotations (HOC). Additionally, 2D joint locations are more heavily affected by camera viewpoint due to perspective projection and occlusions. Therefore, comparing 2D trajectories can be misleading, as these trajectories are affected by nuisance factors such as variations in body shape and camera viewpoint.

Recent advances in deep learning offer a more robust and efficient approach to comparing human movements in video data \cite{Holden2017PhasefunctionedNN, Holden2016ADL}. The key idea is to use a neural network to map the video to a compressed \emph{motion representation} that captures the essence of the movements. This is achieved by using large-scale video data or pose sequences to learn \emph{disentangled motion representations}, \ie, motion representations that are invariant to nuisance factors such as body shape and viewpoint.

For example, \cite{Aberman2019LearningCM} proposes a novel approach for decomposing motion data into dynamic and static representations. Originally developed for motion retargeting, this technique utilizes an encoder-decoder network to separate motion data into skeleton-independent dynamic features and skeleton-dependent static features. The model in~\cite{Coskun2018HumanMA} further decomposes a pose sequence into individual body parts, generating representations for each part separately. This results in motion representations that are suitable for measuring the similarity between different motions of each part. The network is trained with a motion variation loss, enhancing its ability to distinguish even subtly different motions. However, these methods are not directly applicable for distinguishing an individual with ASC as they need very large training datasets to be able to distinguish fine-grained differences in motion, \eg, when an individual is trying to imitate precise movements. A detailed review of related works is provided in Appendix I.

In this work, we propose \method, a novel deep learning-based method to assess motor imitation for distinguishing individuals with ASC in video data.
\method\ uses an encoder-decoder architecture to learn a motion representation or encoding that is disentangled from nuisance factors such as skeletal shape and camera views. Such a disentangled representation is crucial for ensuring that the model accurately captures the essence of the movements themselves, without being influenced by irrelevant variations. For example, people with different body shapes may perform the same movement in slightly different ways due to variations in limb length or body shape. If the model does not disentangle these factors, it might incorrectly attribute these differences to the quality of the imitation rather than as natural variations due to the person's physical characteristics. Similarly, variations in camera angles or distances could make identical motions look different. 
By isolating the motion characteristics from these factors, \method\ can consistently evaluate the quality of motor imitation, regardless of an individual's body shape, or whether the video was recorded from a different angle, thereby eliminating the need for labor-intensive tasks like manual frame-by-frame data cleaning and ad hoc normalization, which are needed in methods like CAMI-3D and CAMI-2D.

To effectively learn these disentangled representations, we employ large-scale synthetic data generated by motion retargeting of virtual characters through the reshuffling of motion, body, and camera views, along with participant data from individuals with ASCs and neurotypical individuals. \method\ automatically assesses a person's imitation performance by computing a similarity score between motion encodings, which can then be used for autism diagnosis.  \method\ addresses critical limitations of existing manual methods (HOC) and automated systems (CAMI-3D and CAMI-2D) by providing a quick, reliable, and easy-to-use tool for assessing imitation. 

The ability of \method\ to precisely quantify motor imitation performance offers a scalable approach for addressing autism heterogeneity in ways that account for a skill, motor imitation, that is fundamental for development of socialization, communication, and other essential skills central to diagnosis of, and targeted intervention for, autism. As such, \method\ has the potential to support more frequent, accessible, and detailed assessments, facilitating earlier and more accurate diagnoses, and more personalized treatment planning.

Specifically, the contributions of this paper are as follows:
\vspace{-.3mm}
\begin{itemize}
    \item \myparagraph{A deep-learning-based approach to motor imitation assessment} 
    \method\ uses an encoder-decoder architecture trained on (a) large-scale synthetic videos generated through motion retargeting and (b) participant videos from individuals with ASCs as well as neurotypical individuals to effectively learn a motion representation that is disentangled from skeletal shape and camera viewpoint, providing a more robust and objective representation for quantitative assessment of motor imitation.

    \item \myparagraph{Interpretability through localized scores} By segmenting the motion representation into different body parts and movement types, \method\ offers localized imitation scores, which not only improve the interpretability of the results but also have the potential to enable tailored interventions based on the specific imitation deficits identified in individuals with ASCs.

    \item \myparagraph{Scalability and practicality} Unlike CAMI-3D, which relies on specialized 3D cameras, hand-crafted normalization, and HOC annotations for training, and CAMI-2D, which also requires hand-crafted normalization and HOC annotations, CAMI-2DNet operates directly on standard video input and requires neither ad hoc normalization nor HOC annotations. This significantly enhances the scalability and practicality of \method\, making it suitable for use in varied settings, including clinics and home environments.

    \item \myparagraph{Empirical validation} 
    Our comparative analysis demonstrates that \method\ strongly correlates with HOC, matches the performance of CAMI-3D, and outperforms both CAMI-2D and HOC in classifying children into diagnostic groups. 
    These results validate \method\ as an effective, practical, and scalable tool for assessing motor imitation in autism diagnosis.

\end{itemize}

\section{Overview of \method}
\label{sec:overview}

\begin{figure*}[!htbp]
    \centering
    \begin{subfigure}{0.345\textwidth}
      \centering
      \includegraphics[width=0.92\linewidth]{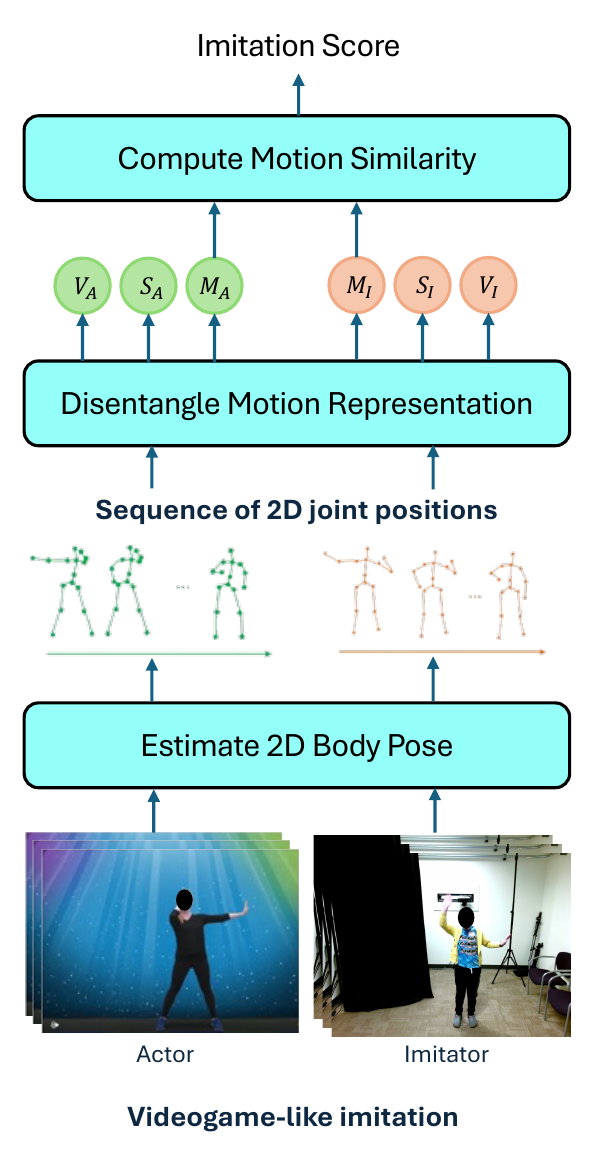}
      \caption{Overview of the \method\ Architecture}
      \label{fig:overview}
    \end{subfigure}
    \hspace{0.01\textwidth}
      \color{gray}
      \rule[5.5ex]{0.05pt}{65ex} 
      \color{black}
    \hspace{0.01\textwidth}
    \begin{subfigure}{0.565\textwidth}
      \centering
      \includegraphics[width=\linewidth]{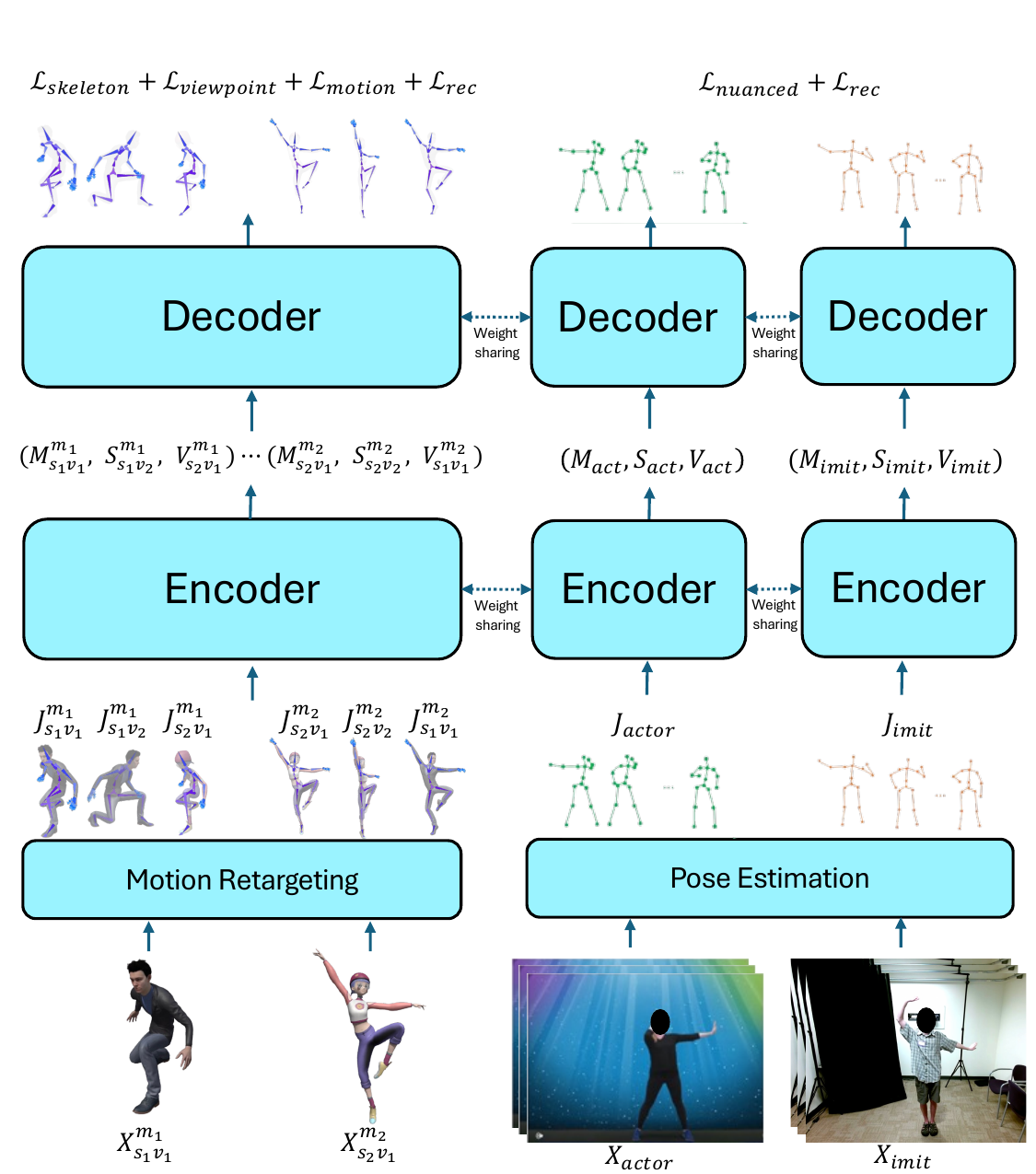}
      \caption{Overview of the \method\ Training Process}
      \label{fig:training_process}
    \end{subfigure}%

    \caption{\textbf{Overview of \method.} 
    (a) Given videos of an actor performing a target action and an individual imitating it, \method\ extracts 2D joint positions using a pose estimation network and encodes the joint trajectories into disentangled motion, shape, and viewpoint components. The imitation score is computed by calculating the cosine similarity of the motion representations ($M_a$ for the actor and $M_i$ for the individual).
    (b) During training, the model learns these disentangled representations from synthetic data generated via motion retargeting (varying motion, shape, and viewpoint) and real participant data from neurotypical individuals and individuals with ASCs. The encoder-decoder architecture is optimized using reconstruction and disentanglement losses, ensuring effective encoding and disentanglement of motion, shape, and viewpoint.}
    
  \label{fig:minds}
\end{figure*}

In this section, we summarize our \method\ method for distinguishing individuals with autism in video. Given a video of an actor performing a sequence of movements and a video of a person imitating these movements, the goal is to produce a score that quantifies how closely the person's movements match those of the actor. This imitation score is then used to help discriminate individuals with ASCs from NT individuals.

\subsection{Stages of \method}
An overview of \method\ is illustrated in Figure~\ref{fig:overview}.
The method comprises three main stages: estimating body pose, learning a disentangled motion representation, and computing the imitation score. Here is a summary of each stage, with further details provided in the corresponding sections.

\begin{itemize}
    \item \mytitle{Estimate 2D Body Pose} Extract a sequence of 2D body joints (e.g., elbows, knees) from each video, converting visual data to 2D trajectories suitable for motion analysis.

    \item \mytitle{Learn a Disentangled Motion Representation} Map the sequence of 2D body joints to a learnable motion representation that is disentangled from nuisance factors such as variations in body shape or camera viewpoint.

    \item \mytitle{Compute Motion Imitation Score} Use the disentangled motion representation to compute a score that quantifies how well a person imitates the movements of the actor.
\end{itemize} 

These stages allow \method\ to yield a more accurate and robust motor imitation assessment by disentangling the dynamics of motion and eliminating distortions caused by irrelevant variables such as body shape and camera viewpoint.

\subsection{Estimating 2D Body Pose}
\label{body-pose-estimation}
The first step in \method\ is to extract the 2D coordinates of the human body joints (\eg, elbows, knees, shoulders) in each video frame, a.k.a. 2D pose estimation. Given a video $\mathbf{X} \in \mathbb{R}^{T \times H \times W \times 3}$, where $T$ represents the number of frames and ($H$, $W$) denotes the height and width of each frame, a pose estimation model predicts the 2D coordinates of key body joints, which we represent as $\mathbf{J} \in \mathbb{R}^{T \times J \times 2}$, where $J$ is the number of body joints. These joint positions form a time series that tracks the subject's body movements throughout the video, providing suitable data for understanding their motion in subsequent stages of motor imitation assessment.

In this work, we utilize a Vision Transformer-based pose estimation model, EViTPose~\cite{Kinfu2023EfficientVT}, due to its ability to capture long-range dependencies between body parts while maintaining computational efficiency. This allows the model to generate accurate joint positions even in complex scenarios, such as when the subject is partially occluded or in varying postures. We use EViTPose as a pre-trained, off-the-shelf pose estimator without any additional fine-tuning.

To further enhance the specificity and interpretability of motor imitation assessments, 
we divide the overall pose sequence $\mathbf{J}$ into $S$ segments, each one corresponding to a specific body part (\eg, arms, legs, torso), and denote the  trajectories of the joints in body part segment $\mathcal{S}$ by $\mathbf{J}_{\mathcal{S}}$. This localization into body parts allows for more detailed insights into which specific areas may be contributing to any observed differences in motor imitation. 
For more details, please refer to Appendix IV-A.

\subsection{Learning Disentangled Representations}
Assessing motor imitation by comparing raw joint trajectories obtained via pose estimation can be misleading because anatomical and viewpoint variations in these trajectories can make it difficult to evaluate accurately how well an individual is imitating a target action. For example, two people performing the same action, such as raising an arm, may appear different due to variations in body posture, like limb length.  Similarly, identical movements can look significantly different when captured from a side view versus a front view.

As discussed in the introduction, methods like CAMI-3D~\cite{Tungen2021ComputerisedAO} attempt to address these variations using hand-crafted techniques, such as adjusting and reorienting joint coordinates to account for changes in body proportions and camera angles, respectively. While effective in controlled settings, these rule-based techniques often fail to generalize to real-world variability. In contrast, \method\ learns disentangled representations of motion, shape, and viewpoint, i.e., it learns motion features that are invariant to body shape or viewpoint variations, as detailed in Section~\ref{sec:learning_disentangled_rep}.

\subsection{Computing Motion Similarity}
After disentangling motion from skeletal and viewpoint variations, \method\ computes an imitation score that quantifies how closely an individual mimics the actor's movements. By comparing the motion representations alone, ignoring irrelevant factors related to body shape and camera viewpoint, \method\ provides a robust and objective measure of motor imitation performance. This score is then used to distinguish typical imitation from potential impairments, such as those seen in ASCs~\cite{Williams2004ASR}. The details of how we compute the motion imitation score are discussed in Section~\ref{sec:computing_imitation_score}.

\section{Learning Disentangled Representations}
\label{sec:learning_disentangled_rep}
As discussed in the previous section, simply relying on raw pose sequences for motor imitation assessment is insufficient due to the entanglement of motion with irrelevant factors like body shape and camera viewpoint. To address these challenges, \method\ automatically learns a motion representation from the raw pose sequences that is disentangled from these nuisance factors. 
In this section, we discuss how \method\ achieves this disentanglement. First, we describe the encoder-decoder model architecture that processes pose sequences to produce disentangled motion, shape, and viewpoint encodings. Next, we discuss the role of training data, specifically motion retargeting and the integration of synthetic and participant data. Finally, we outline the training objectives that guide the model to learn robust and disentangled representations. 
The overall training process is illustrated in Figure~\ref{fig:training_process}, which provides a visual summary of how \method\ leverages both synthetic and real data to achieve effective disentanglement of motion, shape, and viewpoint.

\subsection{Model Architecture}
To effectively learn a representation disentangled from nuisance factors, we employ an encoder-decoder architecture. The encoder compresses the input pose sequences into latent representations that focus on different action components, while the decoder reconstructs the original pose sequence to validate the quality of the learned representation.

\subsubsection{Encoding} 
The encoder is designed to isolate the core characteristics of motion so that the learned representation accurately reflects the subject’s motor abilities, free from distortions caused by body shape and camera perspectives. The encoding process is formally defined as:
\begin{align}
    (\mathbf{M}, \mathbf{S}, \mathbf{V}) = f_{\text{enc}}(\mathbf{J}; \theta_{\text{enc}}),
\end{align}
where $f_{\text{enc}}$ is the encoding network, parameterized by weights $\theta_{\text{enc}}$, which transforms the raw pose sequence $\mathbf{J}$ into three disentangled components: (i) $\mathbf{M}$ is a motion representation that captures the essence of the subject's movement, (ii) $\mathbf{S}$ is a shape representation that models the body shape of the subject, which can vary across individuals, and (iii) $\mathbf{V}$ is a viewpoint representation that accounts for the camera perspective from which the movement is captured.

By disentangling these components, the encoder allows \method\ to focus solely on the core aspects of the motion $\mathbf{M}$, independent of irrelevant factors like body shape $\mathbf{S}$ or camera viewpoint $\mathbf{V}$. This is crucial for enabling accurate motion comparisons across subjects and environments.

\subsubsection{Decoding} 
The decoder plays a vital role in ensuring that the learned latent representation not only captures the essential characteristics of the motion but also retains sufficient information for accurate reconstruction of the original pose sequence. 
The decoder's objective is to reconstruct the pose sequence $\hat{\mathbf{J}}$ from the disentangled representations, $\mathbf{M}$ (motion), $\mathbf{S}$ (shape), and $\mathbf{V}$ (viewpoint). This reconstruction ensures that the latent space adequately represents all the necessary details to model the original movement accurately.
Formally, the decoding process is defined as: 
\begin{align}
    \hat{\mathbf{J}} = f_{\text{dec}}(\mathbf{M}, \mathbf{S}, \mathbf{V}; \theta_{\text{dec}}),
\end{align}
where $f_{\text{dec}}$ is the decoding network, parameterized by weights $\theta_{\text{dec}}$, which maps the disentangled components $\mathbf{M}, \mathbf{S},$ and $\mathbf{V}$ to a reconstructed pose sequence $\hat{\mathbf{J}}$ that matches the original pose sequence $\mathbf{J}$ as closely as possible.
Please refer to Appendix II-A for further details about the encoder-decoder architecture.

\subsection{Training Data}
\subsubsection{Motion Retargeting}
Directly learning disentangled representations from real-world data is inherently challenging due to the absence of explicit information about the underlying motion, body shape, or camera viewpoint. In natural scenarios, the variations in motion are often entangled with differences in body shapes (e.g., height, limb length) and camera perspectives (e.g., angle, distance), making it difficult to isolate the core movement characteristics from irrelevant factors.

As illustrated in Figure~\ref{fig:training_process}, we address this challenge by leveraging \emph{motion retargeting}, which allows us to systematically reshuffle motion, body shape, and camera viewpoint by applying identical movements to different virtual characters with varying body shapes and capturing these motions from multiple viewpoints. For example, as depicted in the bottom left,
the same underlying movement is performed by two distinct virtual characters ($X_{s_1 v_1}^{m_1}$ and $X_{s_2 v_1}^{m_1}$) or viewed from different camera angles ($X_{s_1 v_1}^{m_1}$ and $X_{s_1 v_2}^{m_1}$). This produces a diverse and rich synthetic dataset, allowing the encoder-decoder to effectively disentangle the core motion dynamics from irrelevant factors such as body shape and viewpoint.

We leverage the Synthetic Actors and Real Actions Dataset~\cite{Park2021ABP}, a synthetic dataset generated via motion retargeting. In this dataset, virtual characters from a set $\mathcal{B}$ perform motions from a set $\mathcal{M}$, and these motions are observed from different viewpoints in a set $\mathcal{V}$. Each training sample includes pairs of virtual characters ${s_1, s_2} \subseteq \mathcal{B}$ performing a triplet of motions ${m_1, m_2, m_3} \subseteq \mathcal{M}$, captured from two distinct viewpoints ${v_1, v_2} \subseteq \mathcal{V}$. The motions $m_1$ and $m_2$ are variations within the same motion class -- such as a low jump and a high jump -- while $m_3$ is a distinctly different motion, such as sitting. 
By reshuffling these components, we ensure that the model learns robust, disentangled motion representations that are invariant to skeletal and viewpoint differences, ultimately leading to more accurate and fair motion comparisons.

\subsubsection{Integrating Synthetic and Real Participant Data}
While synthetic data is essential for learning disentangled representations, it introduces a domain gap: synthetic motions are generic and do not fully reflect the specific types of motions we target in motor imitation assessment for distinguishing ASCs. 
To address this gap, we employ a balanced, integrated training strategy, where each training batch includes an equal mix of samples from both synthetic and real participant data. The model is optimized jointly using the corresponding training objectives from both data sources.
Synthetic data lays the foundation for learning robust motion representations by disentangling motion from skeletal and viewpoint variations, while the participant data refines the model's ability to adapt to the target motion types and captures nuanced differences present in practical settings. This combined training approach ensures that the model benefits from the controlled variability of large-scale synthetic data while simultaneously adapting to the complexity and subtleties of real-world scenarios.

\subsection{Training Objectives}
During training, we employ a combination of loss functions tailored to both synthetic and participant data. These loss functions are crucial in guiding the model to effectively separate the core motion from irrelevant factors such as body shape and camera viewpoint, while also capturing the nuanced variations in motor imitation tasks.

For the synthetic data, the model is trained with losses that enforce the disentanglement of motion, shape, and viewpoint, in addition to the reconstruction loss. In contrast, the participant data training utilizes the reconstruction loss alongside a nuanced motion loss, which helps the model capture the subtleties of the motor imitation task.

\subsubsection{Disentanglement Losses}
Here, we describe the disentanglement loss functions, beginning with a common triplet loss formulation that is applied across all action components.

\mytitle{Triplet Loss} 
The triplet loss is designed to ensure that encodings of the same type (motion, shape, or viewpoint) are closer to each other than encodings of different types. The triplet consists of an anchor, a positive example (similar to the anchor), and a negative example (dissimilar to the anchor). The objective of the triplet loss is to minimize the distance between the anchor and the positive example while maximizing the distance between the anchor and the negative example, encouraging separation between distinct factors.
Formally, the triplet loss for an anchor encoding $\mathbf{E}_{\text{A}}$ with its corresponding positive and negative example encodings $\mathbf{E}_{\text{P}}, \mathbf{E}_{\text{N}}$, is~defined~as:
\begin{align}
\!\!
    \mathcal{L}_{\text{triplet}}(\mathbf{E}_{\text{A}}, \mathbf{E}_{\text{P}}, \mathbf{E}_{\text{N}}) = \left[ \| \mathbf{E}_{\text{A}} \!-\! \mathbf{E}_{\text{P}} \|_2^2 - \| \mathbf{E}_{\text{A}} \!-\! \mathbf{E}_{\text{N}} \|_2^2 + \alpha \right]_+\!\! 
    \end{align}
where $\mathbf{E}$ can represent motion, shape, or viewpoint representations, $[ \cdot ]_+$ denotes $\max(0, \cdot)$ and $\alpha$ is a margin parameter that ensures the distance between $\mathbf{E}_{\text{A}}$ and $\mathbf{E}_{\text{P}}$ is smaller than the distance between $\mathbf{E}_{\text{A}}$ and $\mathbf{E}_{\text{N}}$ by at least the margin $\alpha$. 

\mytitle{Shape Disentanglement Loss} To ensure that the shape encoding $\mathbf{S}$ is invariant to variations in motion and viewpoint but still captures differences in body shape, we apply the triplet loss to shape encodings. In each training sample from the synthetic dataset, virtual characters ${s_1, s_2} \subseteq \mathcal{B}$ perform motions ${m_1, m_2, m_3} \subseteq \mathcal{M}$ from two distinct viewpoints ${v_1, v_2} \subseteq \mathcal{V}$. This allows us to create different combinations for anchor, positive, and negative examples. For an anchor shape encoding $\mathbf{S}_{m_1 s_1 v_1}$, where body $s_1$ performs motion $m_1$ from viewpoint $v_1$, the positive example is $\mathbf{S}_{m_2 s_1 v_2}$ (same body, different motion and viewpoint), and the negative example is $\mathbf{S}_{m_1 s_2 v_1}$ (same motion and viewpoint, different body). The shape disentanglement loss is defined as:
\begin{align}
    \mathcal{L}_{\text{shape}} = \mathcal{L}_{\text{triplet}}(\mathbf{S}_{m_1 s_1 v_1}, \mathbf{S}_{m_2 s_1 v_2}, \mathbf{S}_{m_1 s_2 v_1}) .
\end{align}

\mytitle{Viewpoint Disentanglement Loss} Similarly, we apply the triplet loss to disentangle the viewpoint representation $\mathbf{V}$ from the motion and shape. In this case, for an anchor $\mathbf{V}_{m_1 s_1 v_1}$ representing the viewpoint encoding of motion $m_1$ performed by body $s_1$ from viewpoint $v_1$, the positive example is $\mathbf{V}_{m_2 s_2 v_1}$ (different motion and body, same viewpoint), and the negative example is $\mathbf{V}_{m_1 s_1 v_2}$ (same motion and body, different viewpoint). The viewpoint disentanglement loss is:
\begin{align}
    \mathcal{L}_{\text{viewpoint}} = \mathcal{L}_{\text{triplet}}(\mathbf{V}_{m_1 s_1 v_1}, \mathbf{V}_{m_2 s_2 v_1}, \mathbf{V}_{m_1 s_1 v_2}).
\end{align}

\mytitle{Motion Disentanglement Loss} We apply a set of motion-specific loss functions to disentangle the motion representation $\mathbf{M}$ from the shape representation $\mathbf{S}$ and the viewpoint representation $\mathbf{V}$, while capturing both intra-class variations and subtle differences in motor imitation. We employ two motion disentanglement losses: one for training on the synthetic dataset and another for refining the model's performance on real-world data where the differences in motor imitation are more nuanced. For the synthetic dataset, we extend the triplet loss into a quadruplet loss as in~\cite{Park2021ABP}, to ensure that the model is sensitive to small variations within the same motion class. This quadruplet loss introduces a semi-positive example, which represents a variation within the same motion class. 
Given an anchor motion encoding $\mathbf{M}_{m_1 s_1 v_1}$ for motion $m_1$ performed by body $s_1$ from viewpoint $v_1$, a positive example $\mathbf{M}_{m_1 s_2 v_2}$ with the same motion but different body and viewpoint, a semi-positive example $\mathbf{M}_{m_2 s_2 v_2}$ with a variation within the same motion class but different body and viewpoint, and a negative example $\mathbf{M}_{m_3 s_1 v_1}$ with a different motion but same body and viewpoint, the quadruplet loss can formally be defined as:
\begin{equation}
\begin{split}
\!\!\!
\mathcal{L}_{\text{motion}} &= \mathcal{L}_{\text{triplet}}(\mathbf{M}_{m_1 s_1 v_1}, \mathbf{M}_{m_1 s_2 v_2}, \mathbf{M}_{m_3 s_1 v_1}) \\
    &+ \beta \{ \| \mathbf{M}_{m_1 s_1 v_1} - \mathbf{M}_{m_2 s_2 v_2} \|_2 - \gamma \cdot \text{var}(m_1, m_2) \}.\!\!
\end{split}
\end{equation}
The first term is a triplet loss that ensures that the anchor motion remains closer to the positive example than to the negative example. The second term controls the sensitivity of the motion encoding to intra-class variations by penalizing the Euclidean distance between anchor and semi-positive example. The third term penalizes a variation score between the characteristics vectors $v_{m_1}$ and $v_{m_2}$ of the anchor and semi-positive example and is defined as:
\begin{align}
    \text{var}(m_1, m_2) = \frac{\| \mathbf{v}_{m_1} - \mathbf{v}_{m_2} \|_1}{2 \times | \mathbf{v}_{m_1} |}.
\end{align}
These characteristics vectors contain variables such as energy, distance, and height, which influence the shape movement and are provided as metadata in the dataset. Finally, $\beta$ and $\gamma$ are scaling factors that adjust the impact of each term in the loss.

\mytitle{Nuanced Motion Loss} For the participant data, where motor imitation assessments require distinguishing even more subtle differences between a target and an imitated motion,
we introduce a nuanced motion loss to capture these fine distinctions. This loss penalizes the distance between the motion encodings, using the DTW distance between the corresponding pose sequences as a dynamic margin that guides how ``close'' or ``far apart'' these motion encodings should be. 
Given a pair of pose sequences -- $\mathbf{J}_{\text{actor}}$, representing the actor's movements and, $\mathbf{J}_{\text{imit}}$, representing a person's imitated movements -- and their corresponding motion encodings,  $\mathbf{M}_{\text{actor}}$ and $\mathbf{M}_{\text{imit}}$, respectively, the loss is defined as:
\begin{align}
\!\!\!
    \mathcal{L}_{\text{nuanced}} \!=\! 
    \| \mathbf{M}_{\text{actor}} \!-\! \mathbf{M}_{\text{imit}} \|_2^2 + \delta \cdot \text{dist}(\text{DTW}(\mathbf{J}_{\text{actor}}, \mathbf{J}_{\text{imit}})),
    \!\!
\end{align}
where the function $\text{dist}(\text{DTW}(\mathbf{J}_{\text{actor}}, \mathbf{J}_{\text{imit}}))$ calculates the distance between the pose sequences after alignment with Dynamic Time Warping (DTW). The Euclidean distance between the motion encodings $\| \mathbf{M}_{\text{actor}} - \mathbf{M}_{\text{imit}} \|_2^2$ is influenced by this DTW distance, which acts as a margin, and the parameter $\delta$ is a scaling factor that modulates its impact. If the DTW distance between the pose sequences is small (indicating strong alignment in the movements), the encodings are encouraged to be closer together than the ones with a larger DTW distance (indicating less alignment).

\subsubsection{Reconstruction Loss} The reconstruction loss ensures that the latent representations contain sufficient information to accurately reconstruct the original pose sequence. This loss helps maintain the integrity of the learned representation while simultaneously validating the completeness of the disentangled components: motion ($\mathbf{M}$), shape ($\mathbf{S}$), and viewpoint ($\mathbf{V}$). The reconstruction loss is computed as: 
\begin{align}
    \mathcal{L}_{rec}(\mathbf{J}, \mathbf{\hat{J}}) = \frac{1}{T}\frac{1}{J} \sum_{t=1}^T \sum_{j=1}^J \| \mathbf{J}_t^j - \hat{\mathbf{J}}_t^j \|_2^2,
\end{align}
where $\mathbf{J}_t^j$ and $\hat{\mathbf{J}}_t^j$ represent the 2D coordinates of joint $j$ at time $t$ in the original and reconstructed sequences, respectively. 

\subsubsection{Total Loss} 
The total loss integrates the disentanglement, reconstruction, and nuanced motion losses. For the synthetic data, the focus is on disentangling motion, shape, and viewpoint while ensuring that the model can reconstruct the original pose sequences. The total loss for synthetic data is given by: 
\begin{align}
\!\!
    \mathcal{L}_{\text{total-syn}} = \lambda_{\text{dis}} ( \mathcal{L}_{\text{shape}} + \mathcal{L}_{\text{viewpoint}} + \mathcal{L}_{\text{motion}}) + \lambda_{\text{rec}} \mathcal{L}_{rec} ,
    \!\!
\end{align}
where $\lambda_{\text{dis}}$ and $\lambda_{\text{rec}}$ are weighting factors that control the contribution of each component. 

For the participant data, we apply a different combination of losses to ensure that the model can reconstruct real sequences and capture the nuanced differences in motor imitation. The total loss for the participant data is given by: 
\begin{align}
    \mathcal{L}_{\text{total-real}} = \lambda_{\text{rec}}  \mathcal{L}_{\text{rec}}  + \lambda_{\text{nuanced}} \mathcal{L}_{\text{nuanced}} ,
\end{align}
where the weights $\lambda_{\text{rec}}$ and $\lambda_{\text{nuanced}}$ balance the two losses.

During training, the model learns from both synthetic and participant data in a mixed process. The overall total loss is a weighted sum of the synthetic and real data losses:
\begin{align}
    \mathcal{L}_{\text{total}} = \lambda_{\text{syn}}  \mathcal{L}_{\text{total-syn}}  + \lambda_{\text{real}} \mathcal{L}_{\text{total-real}} ,
\end{align}
where $\lambda_{\text{syn}}$ and $\lambda_{\text{real}}$ are weighting factors that balance the contributions of the synthetic and participant data during training. By combining these losses, the model benefits from the strengths of both datasets, leveraging synthetic data for disentanglement and participant data for adapting to the nuanced complexities of real-world motor imitation.

\section{Computing Motion Imitation Score}
\label{sec:computing_imitation_score}
Having learned robust and disentangled representations of motion, shape, and viewpoint during training, the goal of \method\ is to compute an imitation score that quantifies the similarity between the actor's motion and the person's imitated motion. As discussed before, this score is critical for evaluating motor imitation performance for diagnosing ASCs.

To compute this score, \method\ focuses solely on the motion encodings, eliminating the influence of nuisance factors such as differences in body shape  or camera viewpoint. This allows for a more accurate and fair comparison of the movements between the actor and a person. In this section, we detail the process \method\ uses to compute the imitation score, beginning with the encoding of the pose sequences, followed by refining and aligning the motion encodings, and concluding with the calculation of cosine similarity between the actor's and the person's motion encodings.

\subsubsection{Encoding}
Given two pose sequences, $\mathbf{J}_{\text{actor}}$ representing the actor's movements and $\mathbf{J}_{\text{imit}}$ representing the person's imitated movements, we first encode these sequences using the trained encoder $f_{\text{enc}}$, which generates three disentangled components for both the actor and the person: motion encoding $\mathbf{M}$, shape encoding $\mathbf{S}$, and viewpoint encoding $\mathbf{V}$:
\begin{align}
    (\mathbf{M}_{\text{actor}}, \mathbf{S}_{\text{actor}}, \mathbf{V}_{\text{actor}}) &= f_{\text{enc}}(\mathbf{J}_{\text{actor}}; \theta_{\text{enc}}), \\
    (\mathbf{M}_{\text{imit}}, \mathbf{S}_{\text{imit}}, \mathbf{V}_{\text{imit}}) &= f_{\text{enc}}(\mathbf{J}_{\text{imit}}; \theta_{\text{enc}}).
\end{align}

\subsubsection{Optimizing Motion Encodings}
Once the original pose sequences $(\mathbf{J}_{\text{actor}}, \mathbf{J}_{\text{imit}})$ have been encoded, we refine their motion encodings $(\mathbf{M}_{\text{actor}},\mathbf{M}_{\text{imit}})$ to improve the reconstruction of the original pose sequences. The refinement is carried out by minimizing the reconstruction loss $\mathcal{L}_{\text{rec}}$, which measures the difference between the original pose sequences and their reconstructed versions $\hat{\mathbf{J}}_{\text{actor}} = f_{\text{dec}}(\mathbf{M}_{\text{actor}}, \mathbf{S}_{\text{actor}}, \mathbf{V}_{\text{actor}}; \theta_{\text{dec}})$ and $\hat{\mathbf{J}}_{\text{imit}} = f_{\text{dec}}(\mathbf{M}_{\text{imit}}, \mathbf{S}_{\text{imit}}, \mathbf{V}_{\text{imit}}; \theta_{\text{dec}})$, while keeping shape $(\mathbf{S}_{\text{actor}},\mathbf{S}_{\text{imit}})$ and viewpoint $(\mathbf{V}_{\text{actor}}, \mathbf{V}_{\text{imit}})$ encodings and the decoder $f_{\text{rec}}$ freezed. The minimization objective is given by:
\begin{align}
    &\min_{\mathbf{M}_{\text{actor}}} \mathcal{L}_{\text{rec}} ( \mathbf{J}_{\text{actor}}, 
    f_{\text{dec}}(\mathbf{M}_{\text{actor}}, \mathbf{S}_{\text{actor}}, \mathbf{V}_{\text{actor}}; \theta_{\text{dec}})), \\
    &\min_{\mathbf{M}_{\text{imit}}} \mathcal{L}_{\text{rec}} ( \mathbf{J}_{\text{imit}}, 
    f_{\text{dec}}(\mathbf{M}_{\text{imit}}, \mathbf{S}_{\text{imit}}, \mathbf{V}_{\text{imit}}; \theta_{\text{dec}})).
\end{align}

\subsubsection{Computing the Imitation Score}
After optimizing the motion encodings, we ignore the shape and viewpoint encodings and focus solely on comparing the motion representations $\mathbf{M}_{\text{actor}}$ and $\mathbf{M}_{\text{imit}}$. Before computing the similarity between the two motion encodings, we first temporally align the encodings using DTW to account for any differences in timing or duration between the actor’s and the person’s motions.
Following the alignment, we compute the similarity between the motion encodings using cosine similarity, which provides a quantitative measure of how closely the encoded representations of the two motions align. The cosine similarity is given by: 
\begin{align}
    \text{score}(\mathbf{M}_{\text{actor}}, \mathbf{M}_{\text{imit}}) =  \frac{\mathbf{M}_{\text{actor}} \cdot \mathbf{M}_{\text{imit}}}{\| \mathbf{M}_{\text{actor}} \|_2 \| \mathbf{M}_{\text{imit}} \|_2 }.
\end{align}

To enhance the interpretability of the imitation assessment, the final score is computed as a weighted average of the cosine similarity of motion encodings of the actor and an individual for the $\mathcal{S}$ body segments as discussed in Section~\ref{body-pose-estimation}. The final imitation score is computed as:
\begin{align}
    \text{CAMI}(\mathbf{M}_{\text{actor}}, \mathbf{M}_{\text{imit}}) = \sum_{\mathcal{S} \subset \mathcal{J}} w_\mathcal{S} \cdot \left[ \text{score}(\mathbf{M}^\mathcal{S}_{\text{actor}}, \mathbf{M}^\mathcal{S}_{\text{imit}})\right]_+\!\!,
\end{align}
where $w_{\mathcal{S}}$ represents the weight assigned to the body segment ${\mathcal{S}}$, which is a subset of $\mathcal{J}$, the full set of body joint indices. Each body segment ${\mathcal{S}}$ corresponds to a specific set of joint indices (\eg, joints for the left arm or right leg). The weights are determined via cross-validation grid-search hyperparameter tuning. By isolating body segments, \method\ localizes the assessment to specific areas of the body, providing insight into which body part contributes to the imitation differences. For visualization examples of these localized assessments, please refer to Appendix~IV-A.
The final CAMI score ranges between 0 and 1, and quantifies how well the person imitates the actor's movement, with higher values indicating greater similarity and better imitation performance. This score, after normalization using the minimum and maximum values, is then used to discriminate people with ASCs from NT.

\section{Experiments}
In this section, we provide an overview of the synthetic and participant datasets, including details about the participants and experimental procedures. Moreover, we present experiments comparing the effectiveness of our deep-learning-based method, \method, relative to the non-deep-learning methods, CAMI-3D, CAMI-2D, and Human Observation Coding (HOC). The evaluation focuses on construct validity, reliability, and diagnostic classification performance.

\subsection{Dataset details}
\subsubsection{ Synthetic
Actors and Real Actions Dataset}
We employed the synthetic motion dataset SARA~\cite{Park2021ABP}, created using Adobe Mixamo~\cite{mixamo}, to gather sequences of poses from a variety of 3D characters, each with a unique body shape, performing the same motions under kinematic constraints. This dataset comprises motion sequences from 18 different 3D characters across four action categories: Combat, Adventure, Sport, and Dance. Each action sequence comprises a minimum of 32 frames, with a total of 4,428 base motions (\eg, dancing, jumping), having noticeable intra-class variations, resulting in a total of 103,143 variations. Each frame in these sequences contains the 3D coordinates of 17 joints from various body parts, and samples were generated through 2D projection.

\subsubsection{Participant Dataset}
In addition, we incorporated participant data from neurotypical (NT) individuals and individuals with Autism Spectrum Conditions (ASCs) to train and evaluate our method. These participant data were collected as part of a wider-scale study examining imitation skills in autism.
\paragraph{Participants} The participant dataset included 185 people aged 6 to 12 years, comprising 82 children with ASCs and 103 neurotypical (NT) children. We refer to this dataset as CAMI-185.
Among these participants, 47 participants (27 with ASCs, 20 NT) have HOC score annotations, forming a subset we refer to as CAMI-47. See Appendix III-A
for details on the participant demographies and socioeconomic status.
As detailed in~\cite{Tungen2021ComputerisedAO}, and consistent with other prior methods for assessing motor imitation~\cite{Mostofsky2006DevelopmentalDI}, the HOC procedure was designed to derive semi-quantitative scores of motor imitation performance, with summed ratings of individual movements, each scored as inaccurate (0) or accurate (1). Two expert raters, trained by a pediatric neurologist (SHM), scored each participant, with inter-rater and intra-rater reliability established.

The CAMI-47 subset was used to evaluate the performance of the methods against HOC and to train the CAMI-3D method.\footnote{CAMI-2D does not require training the linear regression as it uses the weights learned by CAMI-3D for regressing the CAMI scores.}
The autism diagnoses were based on the Diagnostic and Statistical Manual of Mental Disorders, Fifth Edition (DSM-5)~\cite{Abbas2013DIAGNOSTICAS} criteria as applied by a board-certified Child Neurologist (SHM) with over 30 years of clinical and research experience with children with ASC. Research-reliable assessors confirmed the diagnosis on-site using the Autism Diagnostic Observation Schedule, Second Edition (ADOS-2)~\cite{Mccrimmon2014TestRA}. The parent-report version of the Social Responsiveness Scale, Second Edition (SRS-2)~\cite{Bruni2014TestRS} was also administered. To participate, children needed a full-scale IQ score of at least 80 or at least one index score of 80 (verbal comprehension, visual-spatial, or fluid reasoning index) on the Wechsler Intelligence Scale for Children-Fifth Edition~\cite{Weiss2019WechslerIS}. Additionally, to account for autism-associated differences in general motor abilities, we used the Movement Assessment Battery for Children (mABC), Second Edition~\cite{MovementAB}.
Ethics approval was obtained from the Johns Hopkins University School of Medicine Institutional Review Board before the study began. Written informed consent was obtained from all participants’ legal guardians, and verbal assent was obtained from all children. Recruitment was conducted through local schools and community events. Participants were invited to the Center for Neurodevelopmental and Imaging Research at the Kennedy Krieger Institute for two-day visits and received \$100 compensation for their time.
\paragraph{Procedure}
Children participated in an imitation task involving two movement sequences, (Sequence 1, Sequence 2), each repeated across two trials (Trial A and Trial B). The two sequences included different types of movements (Sequence 1: 14 movements, Sequence 2: 18 movements) that were relatively unfamiliar to the participants (e.g., moving arms up and down like a puppeteer), lacked an end goal, and required the simultaneous movement of multiple limbs. See Appendix III-B for details on the movement types.
The movements were designed to assess imitation of a continuous series of fluid, dynamic movements involving simultaneous articulation of multiple joints. This design provided for assessment of autism-associated difficulties with imitation, in particular challenges with movements requiring dynamic visual-motor integration~\cite{Lidstone2021MovingTU, Gowen2012ImitationIA, MacNeil2012SpecificityOD, McAuliffe2017DyspraxiaIA, Hobson2008DissociableAO, Marsh2013ChildrenWA}.
The stimulus video was displayed on a large TV screen, showing an actor performing dance-like whole-body movements without any background music or sound. The children’s movements were recorded using two Kinect Xbox cameras at 30 frames per second, one positioned in front of the child and the other at the back. 3D data was used for CAMI-3D and video data from the front camera was used for CAMI-2D and \method\ analysis. In the CAMI-185 dataset, 182 participants completed Trial A of both sequences, while fewer participants completed Trial B (Sequence 1: 54, Sequence 2: 61). For the CAMI-47 subset, participants in Trial A were 43 for Sequence 1 and 46 for Sequence 2, while in Trial B they were 40 for Sequence~1 and 36 for Sequence~2.

\begin{figure*}[htbp]
\centering
\begin{subfigure}{.97\textwidth}
  \centering
  \includegraphics[width=\linewidth]{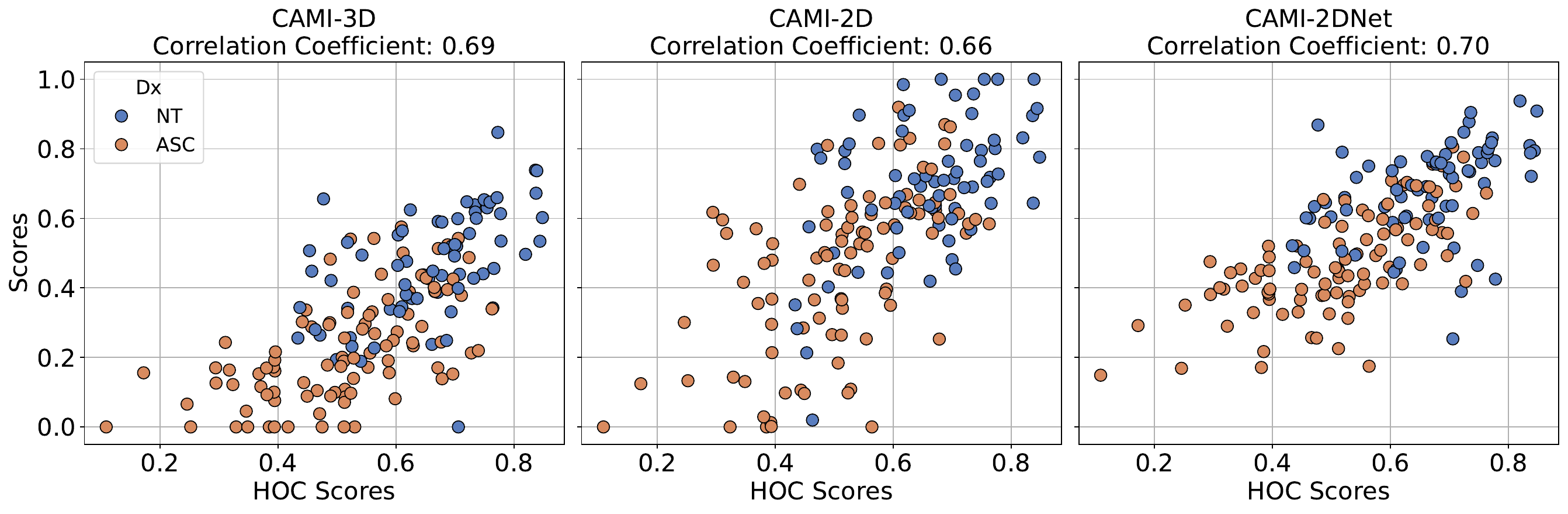}
  \caption{}
  \label{fig:correlation_plot}
\end{subfigure}
\newline
\begin{subfigure}{0.4\textwidth}
  \centering
  \includegraphics[width=\linewidth]{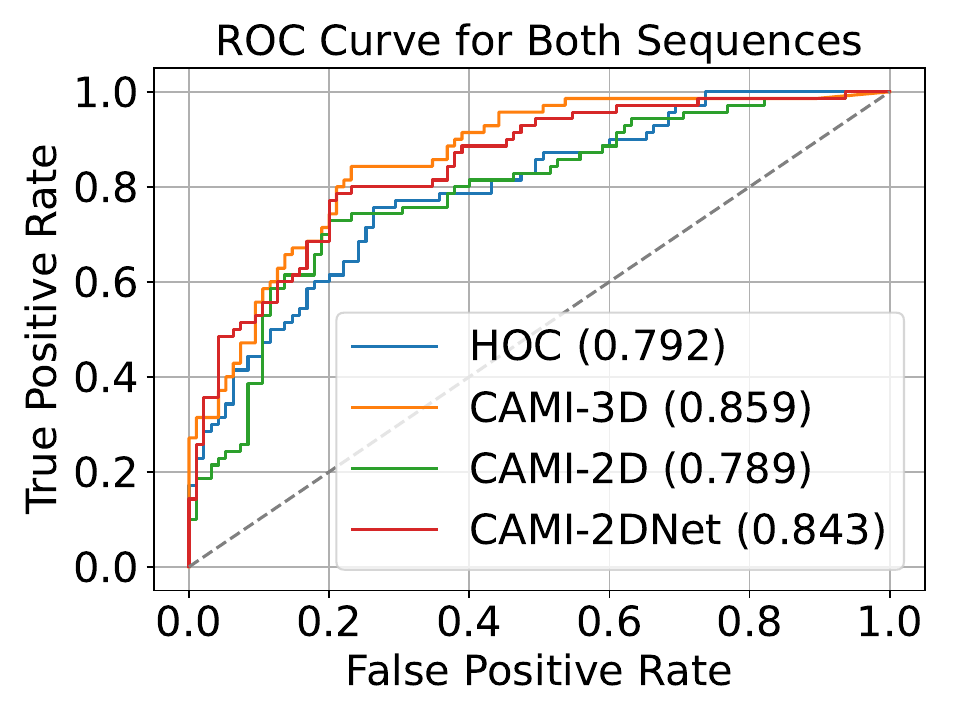}
  \caption{}
  \label{fig:roc_average}
\end{subfigure}%
\begin{subfigure}{0.53\textwidth}
  \centering
  \includegraphics[width=\linewidth]{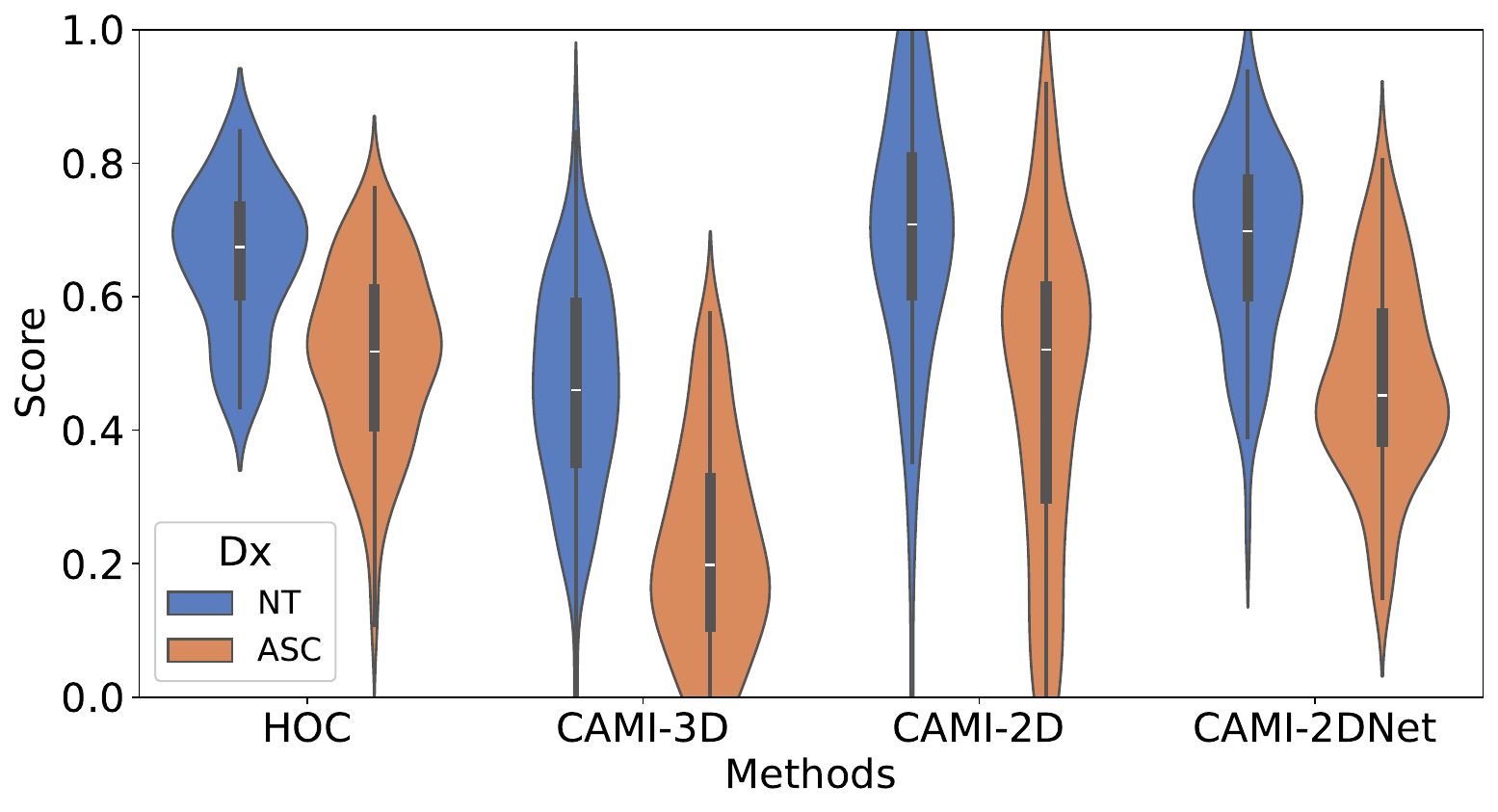}
  \caption{}
  \label{fig:violin_plot}
\end{subfigure}
\caption{Comparing \method, CAMI-2D, CAMI-3D, and human observation coding (HOC) on the CAMI-47 dataset (27 ASCs, 20 NT). 
\textbf{(a) Correlation with HOC Scores}: Scatter plots showing the correlation between HOC scores and the scores from CAMI-3D, CAMI-2D, and \method. 
\method\ has the highest correlation with HOC scores.
\textbf{(b) ROC Curve for Both Sequences}: Receiver operating characteristic (ROC) curve: true positive rate vs. false positive rate as classification threshold is varied. The Area Under the Curve (AUC) indicates the diagnostic ability of the different methods. \method\ (AUC = 0.843) demonstrates comparable performance to CAMI-3D (AUC = 0.859) and superior performance over both HOC (AUC = 0.792) and CAMI-2D (AUC = 0.789). 
\textbf{(c) Violin Plot of Scores}: The violin plots illustrate the distribution of scores for ASC and NT groups across the four methods. \method\ not only shows a clear separation between the ASC groups but also displays less variability within each group, highlighting its robustness and reliability.}
\label{fig:test}
\end{figure*}

\subsection{Results}

\subsubsection{Construct Validity and Test Re-test Reliability}
To verify the construct validity of our method relative to CAMI-3D and CAMI-2D, we analyzed their correlation with the scores obtained from HOC across all sequences and trials of the CAMI-47 dataset. The results, as illustrated in Figure~\ref{fig:correlation_plot}, show strong positive correlations between the three methods and HOC. Specifically, the correlation coefficients were 0.69 for CAMI-3D, 0.66 for CAMI-2D, and 0.70 for \method. Notably, \method, which operates entirely without supervision from HOC, demonstrated the highest correlation with HOC scores. This strong correlation not only highlights the accuracy and reliability of \method\ but also underscores its potential as a highly effective tool for analyzing the participant data independently of HOC supervision. 
Despite strong correlation, indicating similar coarse-level assessment, \method\ and HOC differ in diagnostic discrimination: HOC’s subjective binary scoring can overlook subtle variations in imitation, whereas \method’s fine-grained features capture nuanced patterns, yielding superior diagnostic discrimination (see next).

\begin{figure*}[htbp]
\centering
\begin{subfigure}{0.248\textwidth}
  \centering
  \includegraphics[width=\linewidth]{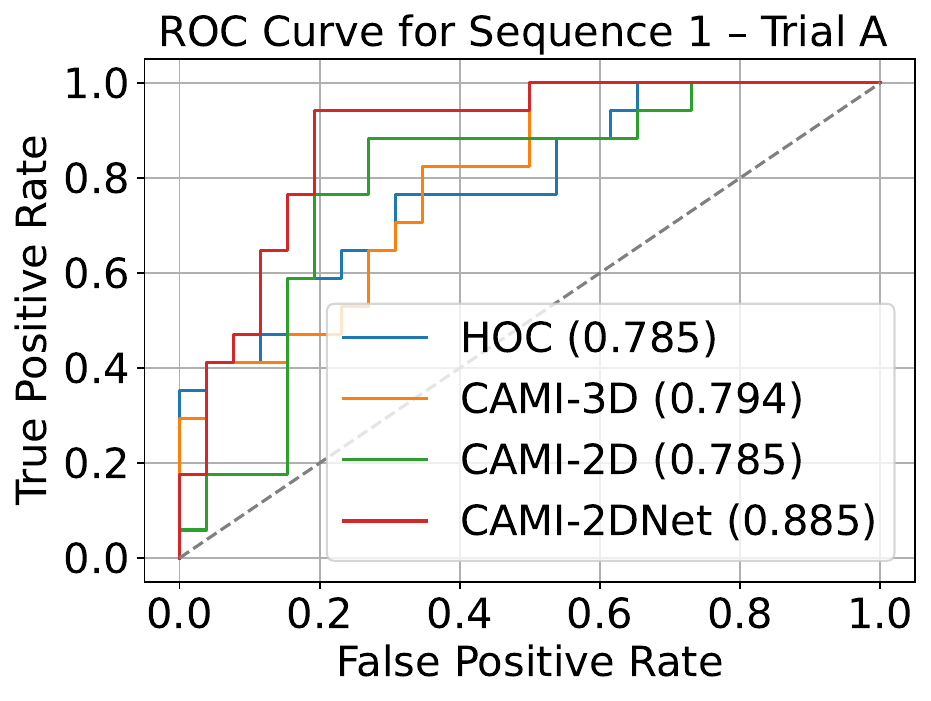}
  \caption{}
  \label{fig:seq_1_trial_a}
\end{subfigure}%
\begin{subfigure}{0.248\textwidth}
  \centering
  \includegraphics[width=\linewidth]{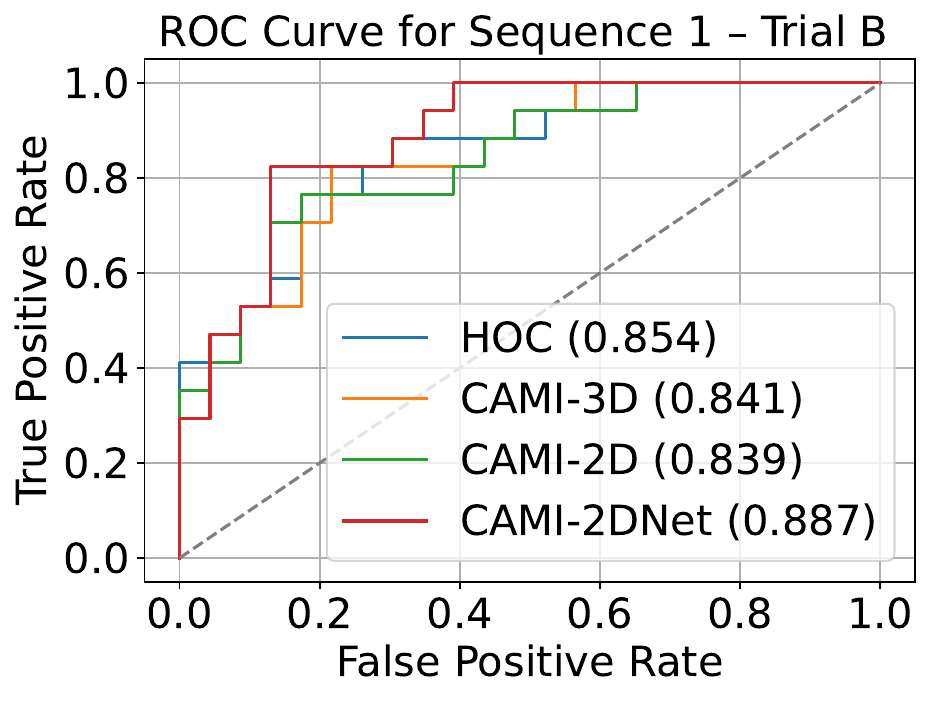}
  \caption{}
  \label{fig:seq_1_trial_b}
\end{subfigure}
\begin{subfigure}{0.248\textwidth}
  \centering
  \includegraphics[width=\linewidth]{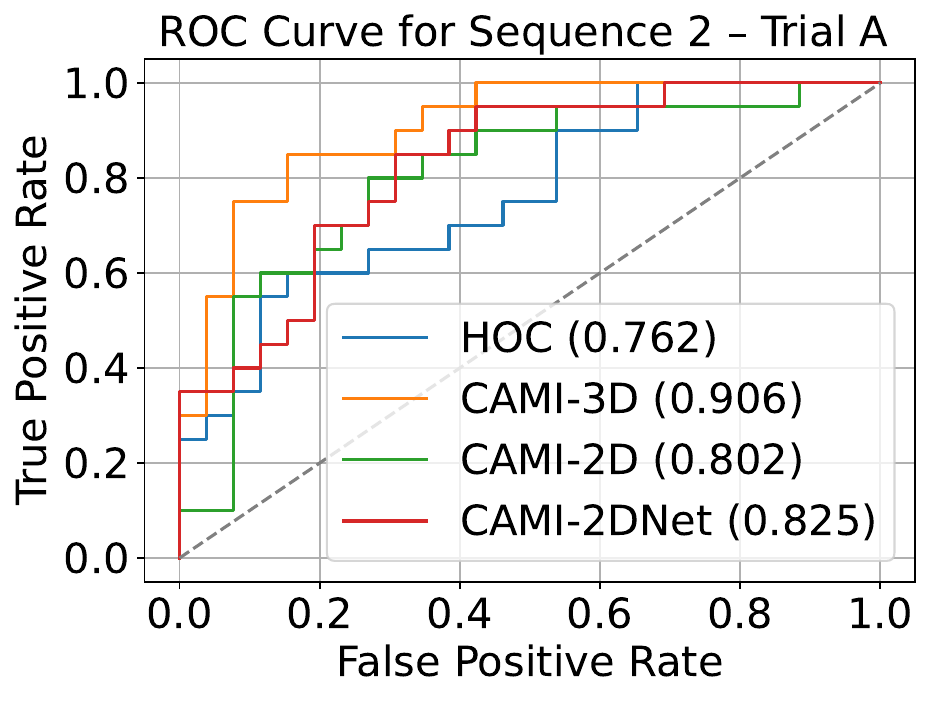}
  \caption{}
  \label{fig:seq_2_trial_a}
\end{subfigure}%
\begin{subfigure}{0.248\textwidth}
  \centering
  \includegraphics[width=\linewidth]{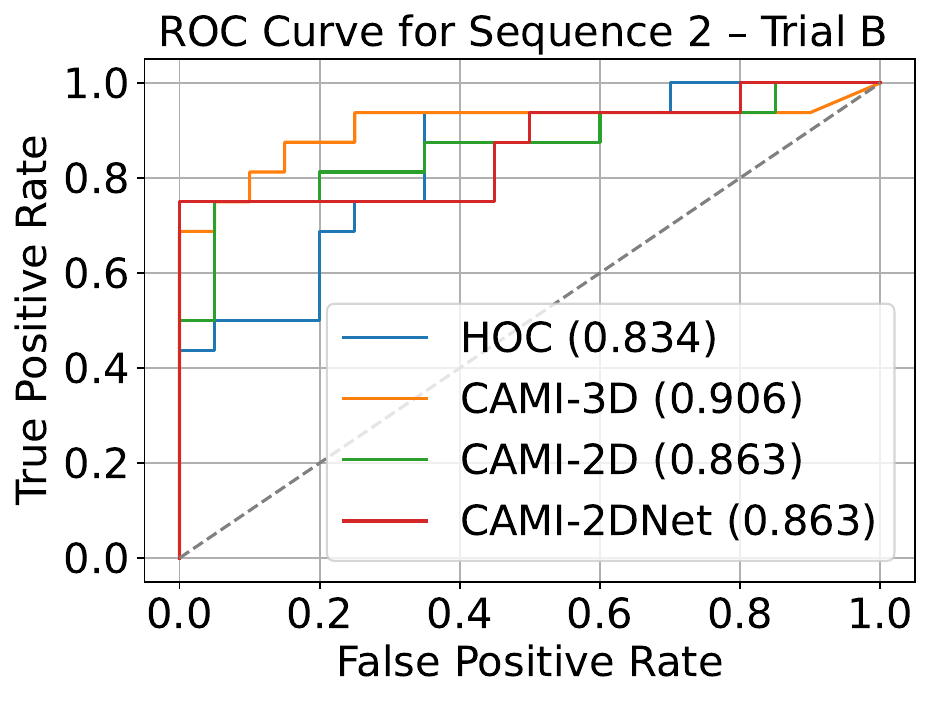}
  \caption{}
  \label{fig:seq_2_trial_b}
\end{subfigure}
\newline
\begin{subfigure}{0.248\textwidth}
  \centering
  \includegraphics[width=\linewidth]{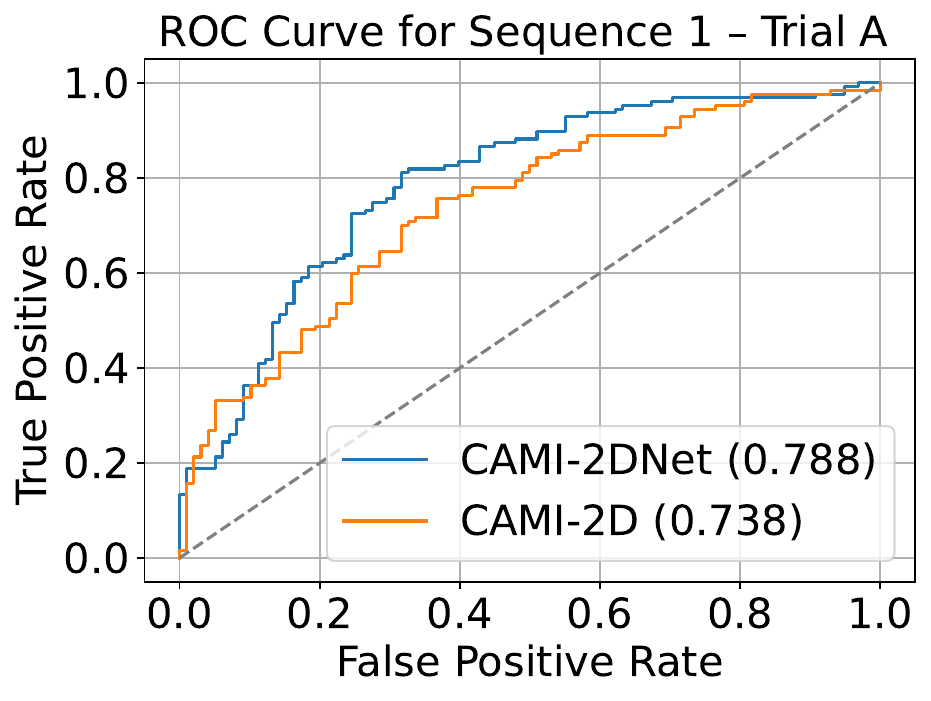}
  \caption{}
  \label{fig:all_seq_1_trial_a}
\end{subfigure}%
\begin{subfigure}{0.248\textwidth}
  \centering
  \includegraphics[width=\linewidth]{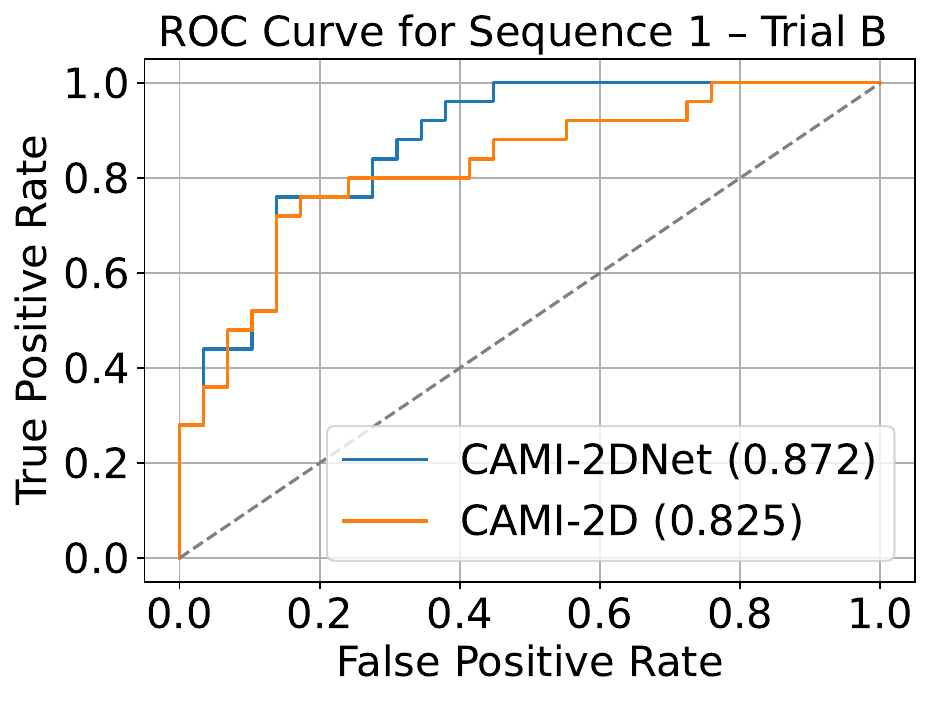}
  \caption{}
  \label{fig:all_seq_1_trial_b}
\end{subfigure}
\begin{subfigure}{0.248\textwidth}
  \centering
  \includegraphics[width=\linewidth]{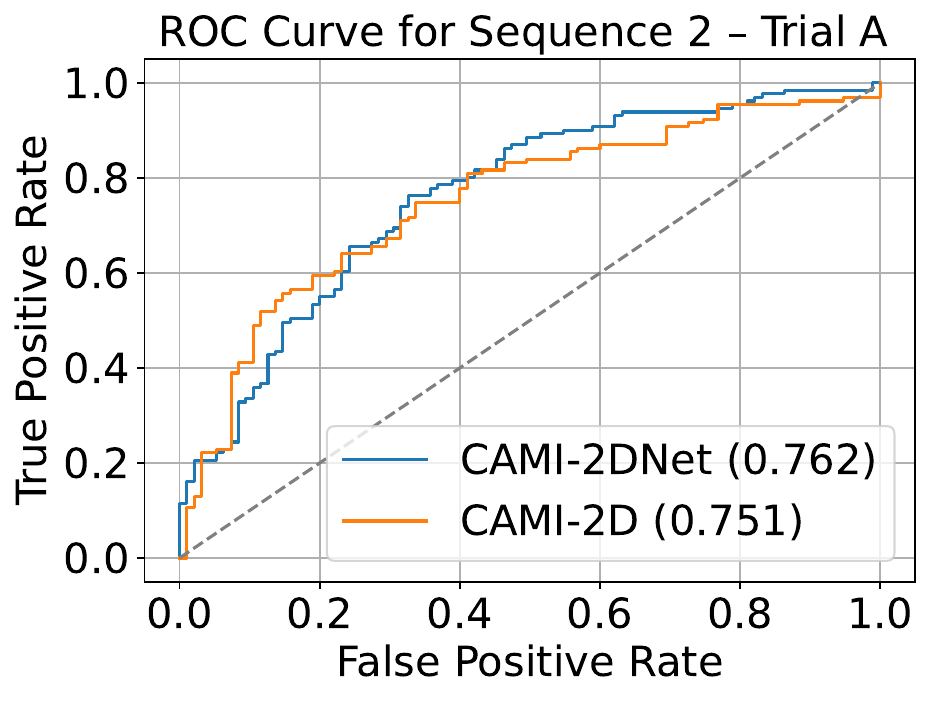}
  \caption{}
  \label{fig:all_seq_2_trial_a}
\end{subfigure}%
\begin{subfigure}{0.248\textwidth}
  \centering
  \includegraphics[width=\linewidth]{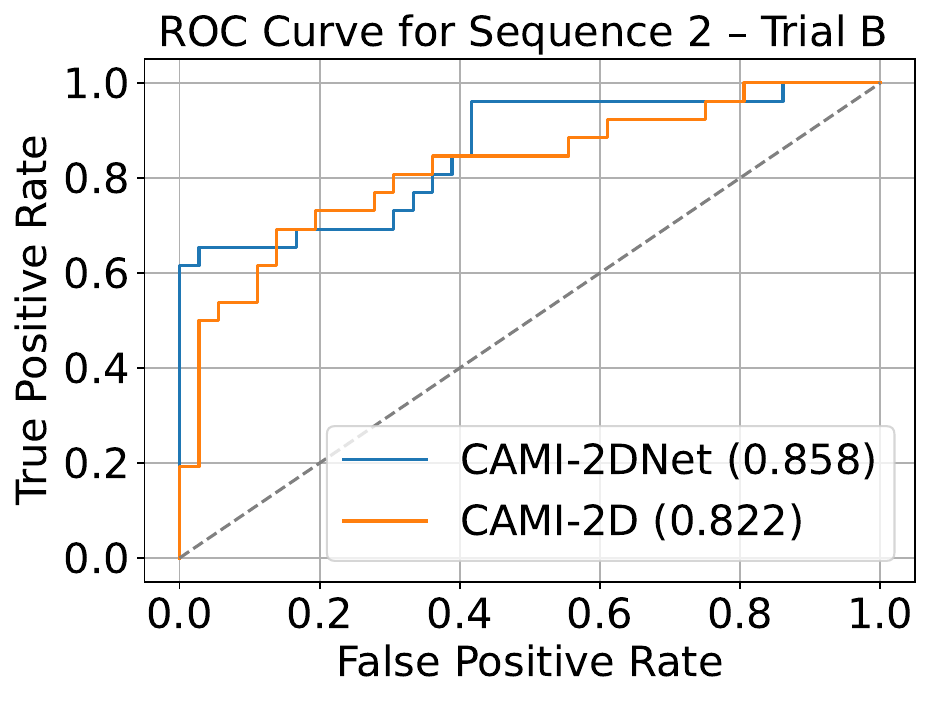}
  \caption{}
  \label{fig:all_seq_2_trial_b}
\end{subfigure}
\caption{\textbf{Receiver Operating Characteristic (ROC) curves comparing the diagnostic performance of HOC, CAMI-3D, CAMI-2D, and \method\ across two datasets: CAMI-47 and CAMI-185.} 
The top row (a-d) presents results on the CAMI-47 dataset for two sequences, each consisting of two trials. \method\ consistently outperforms HOC and CAMI-2D and demonstrates comparable or superior performance to CAMI-3D. 
The bottom row (e-h) shows results on the CAMI-185 dataset, comparing \method\ with CAMI-2D across two sequences and two trials. \method\ achieves higher diagnostic accuracy in all trials, demonstrating a higher AUC than CAMI-2D.
} 
\label{fig:roc_each}
\end{figure*}

\subsubsection{Diagnostic Classification Ability}
We evaluated the performance of our method, \method, relative to CAMI-3D, CAMI-2D, and HOC in classifying children into diagnostic groups by computing the receiver-operating characteristic (ROC) curve across all sequences and trials of the CAMI-47 dataset. Larger areas under the curve (AUC) indicate better discriminative ability, as shown in Figure~\ref{fig:roc_average}.
CAMI-3D demonstrated the highest performance with an AUC of 0.859, indicating its superior capability in distinguishing between diagnostic groups. The 3D nature of this method likely contributes to its better performance.
The CAMI-2D method, which operates on 2D video data, showed an AUC of 0.789. While its performance is slightly lower than that of CAMI-3D, it provides valuable discriminative ability, comparable to the HOC method (AUC = 0.792). The reliance on 2D data without leveraging the depth information of 3D data might account for this difference.
\method\ achieved an AUC of 0.843, demonstrating comparable performance to CAMI-3D and superior performance over both CAMI-2D and HOC. \method's ability to operate directly on video data without requiring HOC annotations during training offers a significant practical advantage. This makes \method\ a highly effective and efficient tool for diagnostic classification, balancing high performance with operational simplicity. 
Furthermore, the violin plots, in Figure~\ref{fig:violin_plot}, showing the distribution of scores for the NT and ASC groups across the four methods, demonstrate that \method\ not only exhibits a distinct separation between ASCs and NT groups but also shows relatively reduced variability within each group, underscoring its robustness and reliability.

The ROC curve for both sequences, consisting of two trials in CAMI-47, is shown in Figure~\ref{fig:roc_each} (a-d). For instance, in both trials of Sequence 1, \method\ achieved the highest AUC (Trial A: 0.885, Trial B: 0.887), outperforming CAMI-3D (Trial A: 0.794, Trial B: 0.84), CAMI-2D (Trial A: 0.785, Trial B: 0.839, and HOC (Trial A: 0.785, Trial B: 0.854).  
For Sequence 2 - Trial A, \method\ achieved an AUC of 0.825, which is higher than CAMI-2D (0.802)\footnote{The results for CAMI-2D reported in this paper differ from those reported in the original CAMI-2D paper~\cite{Lidstone2021AutomatedAS} due to two main reasons: i) the current evaluation included 46 participants, compared to the 40 participants in the original study, and ii) the analysis in this paper employs cross-validation, whereas the original paper did not use cross-validation in its evaluation. These factors contribute to variations in the performance outcomes observed.} and HOC (0.762), but lower than CAMI-3D (0.906). In Sequence 2 - Trial B, \method\ and CAMI-2D both achieved an AUC of 0.863, while CAMI-3D scored the highest at 0.906 and HOC obtained 0.834.
These trials demonstrate the consistent and high performance of \method\ across different sequences and trials, reinforcing its capability as a practical and effective tool for diagnostic classification. 

The ROC curves across both sequences and trials in the CAMI-185 dataset further highlight \method's consistent superiority compared to CAMI-2D. As shown in Figure~\ref{fig:roc_each} (e-h), \method\ achieves higher AUC scores in both trials of Sequence 1 (Trial A: 0.787 vs. 0.737, Trial B: 0.853 vs. 0.824) and Sequence 2 (Trial A: 0.767 vs. 0.751, Trial B: 0.856 vs. 0.824).
Overall, \method\ outperforms CAMI-2D and HOC in terms of diagnostic ability and maintains a strong correlation with HOC scores. Moreover, \method\ performs comparably to CAMI-3D while offering greater practicality by operating directly on video data and without needing labor-intensive HOC annotations and ad hoc normalization steps.  

\medskip
\medskip

\section{Limitations and future directions}
While \method\ shows strong promise as a practical and scalable tool for motor imitation assessment, it has some limitations that open avenues for future research. First, learning disentangled motion representations is based on synthetic data, which may not fully capture the variability and complexity of real-world human movements. To bridge this gap, we employed integrated training with real participant data, but future work should explore other ways, such as domain adaptation. Second, deploying \method\ in real-world clinical or mobile settings introduces challenges related to computational efficiency and hardware limitations, and comprehensive clinical assessment. Optimizing the model for lightweight inference is a critical next step toward broader accessibility and scalability.
It should also be noted that motor imitation difficulties captured by CAMI-2DNet represent only one aspect of autism. Thus, CAMI-2DNet can be a useful tool that complements, but does not necessarily replace, other diagnostic and treatment assessments that tap into autism diagnostic criteria, such as social-communicative difficulties and restricted interests and repetitive behaviors.
Finally, motor imitation behaviors could be influenced by cultural and environmental factors. Expanding training and evaluation datasets to include diverse populations would help improve the model's robustness and fairness across different demographic and cultural contexts. 

\section{Conclusion}
We introduced \method, a scalable and interpretable deep learning-based approach to motor imitation assessment in video data. \method\ uses 2D pose estimation techniques to extract 2D joint trajectories from the video. These trajectories are then mapped to a motion representation that is disentangled from nuisance factors such as body shape and camera viewpoint. A motor imitation score is then computed by comparing the motion representation of an individual to that of the actor. Our experiments demonstrate that \method\ performs on par with CAMI-3D in discriminating ASC vs neurotypical children, and outperforms both HOC and CAMI-2D, while offering greater practicality by operating directly on video data and without the need for ad hoc data normalization and HOC annotations. These results highlight \method\ as an effective and accessible tool for assessing motor imitation in children with ASCs and related developmental conditions.

\bibliography{IEEEabrv, bib.bib}{}
\bibliographystyle{IEEEtran}

\clearpage
\appendices

\re{
\section{Related Works}
\label{sec:related_works}
In this section, we review prior research on motion similarity assessment, autism-related motion analysis, and motion retargeting.
\subsection{Motion Similarity Assessment Methods}
Motion similarity assessment in computer vision focuses on evaluating how closely the actions of an agent resemble reference actions, both in terms of precision and overall quality. This problem has been studied extensively across domains such as human–computer interaction, sports training, rehabilitation, and performance evaluation. Over time, the field has evolved from handcrafted features and classical machine learning methods toward data-driven deep learning approaches that leverage large-scale datasets and powerful pose estimation models.

\myparagraph{Handcrafted Features and Classical Models} The earliest methods relied heavily on handcrafted spatiotemporal features and classical statistical models. For instance, the earliest methods~\cite{Michelet2012AutomaticIA, Schmidt2012MeasuringTD, Paxton2012FramedifferencingMF, Chaudhry2009HistogramsOO, Ravichandran2009ViewinvariantDT} employed descriptors like Histogram of Oriented Gradients (HOG) and Histogram of Optical Flow (HOF)~\cite{Laptev2005OnSI} to capture motion information. These descriptors were often aggregated into Bag-of-Words (BoW) representations~\cite{FeiFei2005ABH}.

The similarity between these BoW representations was then measured using Dynamic Time Warping (DTW)~\cite{Bellman1959OnAC}, providing a measure of imitation accuracy. 

Although these approaches were foundational, they often struggled with the variability and complexity of human actions because they relied on predefined features that might not fully capture the subtle dynamics of movements. 
In addition to these traditional methods, motion capture (MoCap) emerged as an alternative for similarity assessment. These systems provide precise and high-dimensional motion data, enabling more accurate analysis of the subtleties in human movement. 

\myparagraph{Pose Estimation and Skeleton-Based Representations} The advent of deep-learning-based pose estimation marked a major turning point in motion similarity research. Modern pose estimation models can detect and track 2D or 3D skeletal joints directly from RGB video, eliminating the need for expensive MoCap equipment while providing interpretable, high-level motion descriptors. This greatly increased the accessibility and scalability of motion analysis, enabling applications in unconstrained real-world settings~\cite{Jammalamadaka2012VideoRB, Kim2018RealtimeDE}. 
For instance, \cite{Jammalamadaka2012VideoRB}, a highly compact feature vector is constructed by utilizing the angles of various body parts from a pose sequence, enabling the measurement of distances between poses. Similarly, in \cite{Kim2018RealtimeDE}, the similarity between two dance poses is assessed by analyzing the joint angles of the individual in each frame. 

\myparagraph{Deep Learning for Spatiotemporal Representation Learning} Beyond handcrafted encodings, deep neural networks have enabled data-driven representation learning for motion similarity. 3D Convolutional Neural Networks (CNNs), have been crucial in extracting and learning detailed spatiotemporal features from video data or pose sequences. Specifically, convolutional autoencoders can autonomously learn a manifold to represent motion. These advanced techniques enable effective learning from large datasets, facilitating more accurate and robust motion representations~\cite{Holden2017PhasefunctionedNN, Holden2016ADL}. 
Aristidou~\etal~\cite{Aristidou2018DeepMA}, introduced the concept of motion words where motion sequences are divided into short-term movements, which are then clustered in high-dimensional feature space to serve as motion representations. 

Recent works have also explored factorized representations. For example, Aberman~\etal~\cite{Aberman2019LearningCM}  proposed decomposing motion data into dynamic and static features. Originally developed for motion retargeting, this technique utilizes an encoder-decoder network to separate motion data into skeleton-independent dynamic features and skeleton-dependent static features. 
Building on this, Park~\etal~\cite{Park2021ABP} further decomposes a pose sequence into individual body parts, generating encodings for each part separately. This results in motion encodings that are suitable for measuring the similarity between different motions of each part. The network is trained with a motion variation loss, enhancing its ability to distinguish even subtly different motions.

\subsection{Autism-Related Motion Assessment}

Motor impairments, ranging from atypical posture to difficulties in fine motor coordination and imitation, are commonly reported in Autism Spectrum Disorder (ASD)~\cite{Bhat2020MotorII, cook2013atypical}. These impairments serve as diagnostic markers and targets for intervention. Consequently, quantitative motion assessment has received increasing attention in autism research ~\cite{Edwards2014AMO, fournier2010motor}

\myparagraph{Clinical and Observational Approaches} Standard approaches to motor assessment in ASD have relied primarily on clinical observation and standardized scales, such as the Human Observation Coding (HOC) based on the Movement Assessment Battery for Children (MABC) and the Bruininks–Oseretsky Test of Motor Proficiency (BOT-2) ~\cite{MovementAB, bruininks1978}. While informative, such methods are labor-intensive, subject to inter-rater variability, and often lack the granularity needed to capture subtle atypicalities in motor patterns~\cite{Edwards2014AMO}.

\myparagraph{Automated Approaches} To address these limitations, researchers have adopted objective measurement technologies. Motion capture systems and inertial measurement units (IMUs) have been employed to quantify gait parameters, gesture kinematics, and imitation accuracy in ASD populations~\cite{ de2010conversational, Tungen2021ComputerisedAO, marsh2013autism}. 
Recent progress in pose-estimation frameworks has enabled markerless tracking of joint trajectories from video, offering a scalable alternative to mocap-based systems~\cite{Lidstone2021AutomatedAS, kojovic2021using}.

\subsection{Motion Retargeting Methods}
Motion retargeting refers to the process of transferring motion data from one character or skeleton to another, often differing in body shape, proportions, or kinematic constraints. The goal is to preserve the dynamics of the source motion while adapting it naturally to the target. Retargeting is a core task in computer graphics, gaming, and animation pipelines, where reusable motion libraries significantly reduce the time and cost of character animation. Beyond entertainment, retargeting is increasingly used in robotics, VR/AR, and rehabilitation, where demonstrations must adapt across diverse bodies or embodiments.

Early motion retargeting relied primarily on kinematic mapping. Source joint trajectories were geometrically mapped to the target skeleton, and inverse kinematics (IK) was applied to enforce constraints such as joint limits and end-effector placement~\cite{gleicher1998retargetting, savenko2002using}. 
While conceptually straightforward, these methods struggled to handle large differences in body proportions and often produced unnatural artifacts such as foot sliding or distorted limb trajectories. Physics-based retargeting methods attempted to overcome these limitations by incorporating dynamics into the optimization, ensuring that retargeted motions remained physically plausible~\cite{tak2005physically, komura2000creating}. 

Commercial tools have made motion retargeting more accessible to practitioners and artists. Adobe Mixamo~\cite{mixamo}, for instance, offers an online platform with automatic rigging and motion retargeting capabilities. Users can upload custom characters and instantly apply animations from Mixamo’s library, enabling rapid prototyping for games and VR experiences. Similarly, Blender~\cite{blender}, a widely used open-source 3D creation suite, provides built-in motion retargeting through its armature and constraint systems. Blender users can transfer animations between characters with different skeletons, fine-tuning the process with IK solvers and animation layers. 

Recent advances in deep learning have transformed motion retargeting by enabling data-driven representations that capture motion semantics. Neural networks can disentangle motion into invariant dynamic components (e.g., walking style) and skeleton-dependent static components (e.g., limb lengths or proportions), facilitating transfer across diverse characters\cite{zhang2024semantics, ye2024skinned}.

Although primarily developed for animation, motion retargeting has strong conceptual ties to motion similarity assessment. Both tasks require representations that capture the essence of motion while being invariant to irrelevant factors (e.g., body shape, viewpoint). Techniques developed for retargeting, such as motion decomposition, are important in similarity learning. Conversely, similarity metrics can serve as objective evaluation criteria for retargeted motions, ensuring that stylistic and dynamic fidelity are preserved across skeletons.
}

\section{Network Design and Training Configuration}
\subsection{Model Architecture Details}
\label{sec:model_architecture}
The encoder-decoder architecture we employed consists of three encoders -- Motion Encoder, Body Encoder, and View Encoder -- along with a Decoder module as in~\cite{Aberman2019LearningCM, Park2021ABP}. Each encoder disentangles a specific aspect of the input pose sequences: motion dynamics, body shape, and camera viewpoint, respectively. The decoder reconstructs the input 2D pose sequence, ensuring that the disentangled representations retain sufficient information to reconstruct the original poses.

\paragraph{Motion Encoder}
The Motion Encoder processes the temporal dynamics of input pose sequences. It uses three convolutional layers, each with a kernel size $k=8$ and a stride $s=2$, followed by leaky ReLU (LReLU) activation. All convolutional layers use reflected padding. The input has $(2\!\times\!\! J_{\mathcal{S}})$ channels, where $J_{\mathcal{S}}$ represents the number of joints of a body segment $\mathcal{S}$. The number of channels progressively increases to 128, allowing the encoder to extract higher-level motion features at multiple resolutions.

\paragraph{Body Encoder}
The Body Encoder disentangles body shape information. It uses convolutional layers with a smaller kernel size $k=7$ and stride $s=1$, followed by max pooling (MP) to downsample spatial features. The third convolutional layer incorporates global max pooling to capture global body shape features. Finally, a $1 \times 1$ convolution reduces the dimensionality of the feature maps to 16 channels.

\paragraph{View Encoder}
The View Encoder extracts viewpoint-related features, disentangling variations due to camera angles. Similar to the Body Encoder, it applies convolutional layers with $k=7$, followed by average pooling (AP). The third convolutional layer uses global average pooling to summarize viewpoint information globally, and a final $1 \times 1$ convolution reduces the output to 8 channels.

\paragraph{Decoder}
The Decoder reconstructs the input 2D joint locations from the disentangled motion, skeletal, and viewpoint representations. It consists of three layers that progressively upsample the features. Each layer includes an upsampling operation followed by convolution, dropout, and LReLU activation. Dropout is applied in the first two layers to reduce overfitting and improve generalization. The final layer outputs $2\!\times\!\! J_{\mathcal{S}}$ channels, corresponding to the 2D locations of the $J_{\mathcal{S}}$ joints for a body segment $\mathcal{S}$. 

Table~\ref{tab:network_architecture} provides a detailed summary of the encoder-decoder architecture, including the configuration of each layer and the dimensionality of the feature maps.

\subsection{Implementation Details}
\revision{
Our model is implemented using the PyTorch framework. Training is conducted on NVIDIA RTX A5000 GPUs with 24 GB memory. We utilize the Adam optimizer with parameters $\beta_1 = 0.9$, $\beta_2 = 0.999$, and $\epsilon = 10^{-8}$. L2 regularization with a weight decay of 0.01 is applied to prevent overfitting. The initial learning rate is set to $10^{-3}$, and we employ an exponential decay schedule with a decay rate of 0.98 every one-third of an epoch. The model is trained for 210 epochs using a batch size of 2,048 and 12 data loader workers to facilitate efficient data loading. For data preprocessing, 2D pose sequences are normalized to have zero mean and unit variance. Data augmentation techniques, including random horizontal flipping and temporal jittering, are applied to enhance the robustness of the model.
In our experiments, we use the following loss weight coefficients: $\alpha = 0.3$, $\beta = 1.0$, $\gamma = 0.2$, $\delta = 0.02$, $\lambda_{\text{dis}} = 0.7$, $\lambda_{\text{rec}} = 1.0$, $\lambda_{\text{nuanced}} = 1.0$, $\lambda_{\text{syn}} = 1.0$, and $\lambda_{\text{real}} = 1.0$.
}
\begin{table}
\centering 
\caption{\textbf{Network Architecture Summary.} The table describes the network structure of the encoder-decoder architecture for disentangling motion, body shape, and viewpoint representations. Conv., LReLU, MP, AP, Upsample, and Dropout denote convolution, leaky ReLU, max pooling, average pooling, upsampling, and dropout layers, respectively. 
The parameters $k$ and $s$ represent the kernel width and stride, respectively, and the rightmost column reports the number of input and output channels for each layer.}
\label{tab:network_architecture} 
\begin{tabular}{|@{\;}l@{\;}|@{\;}l@{\;}|@{\;\;}c@{\;\;}|@{\;\;}c@{\;\;}|@{\;}c@{\;}|} 
\hline 
\textbf{Name} & \textbf{Layers} & $k$ & $s$ & \textbf{in/out} \\ 
\hline 
\multirow{3}{*}{\shortstack{Motion\\Encoder}} & Conv. + LReLU & 8 & 2 & ($2\!\times\!\! J_{\mathcal{S}}$)/64 \\
& Conv. + LReLU & 8 & 2 & 64/96 \\
& Conv. + LReLU & 8 & 2 & 96/128 \\ 
\hline 
\multirow{4}{*}{\shortstack{Body\\Encoder}} & Conv. + LReLU + Max Pooling (MP) & 7 & 1 & ($2\!\times\!\! J_{\mathcal{S}}$)/32 \\
& Conv. + LReLU + MP & 7 & 1 & 32/48 \\
& Conv. + LReLU + Global MP & 7 & 1 & 48/64 \\
& Conv. & 1 & 1 & 64/16 \\ 
\hline 
\multirow{4}{*}{\shortstack{View\\Encoder}} & Conv. + LReLU + Average Pooling (AP) & 7 & 1 & ($2\!\times\!\! J_{\mathcal{S}}$)/32 \\
& Conv. + LReLU + AP & 7 & 1 & 32/48 \\
& Conv. + LReLU + Global AP & 7 & 1 & 48/64 \\
& Conv. & 1 & 1 & 64/8 \\
\hline 
\multirow{3}{*}{Decoder} & Upsample + Conv. + Dropout + LReLU & 7 & 1 & 152/128 \\
& Upsample + Conv. + Dropout + LReLU & 7 & 1 & 128/64 \\
& Upsample + Conv. & 7 & 1 & 64/($2\!\times\!\! J_{\mathcal{S}}$) \\ 
\hline 
\end{tabular} 
\end{table}

\section{Participant Demographics and Movement Types}
\revision{
\subsection{Participant Demographics}
\label{sec:participant_demographics}
Participant demographics and family socioeconomic status are summarized in Table~\ref{tab:participant_demographics}. The Autism Spectrum Condition (ASC) group consisted of 81 participants (70 males, 11 females) with a mean age of 10.3 years (SD = 1.7), while the neurotypical (NT) group included 104 participants (74 males, 30 females) with a mean age of 10.2 years (SD = 1.4). The racial composition of the ASC group was primarily White (n = 58), followed by African American (n = 12), Multiracial (n = 10), Asian (n = 1), and no participants with unknown racial identity. The NT group displayed a similar distribution, with the majority identifying as White (n = 82), followed by African American (n = 8), Asian (n = 6), and Multiracial (n = 8).

With respect to ethnicity, 11 participants in the ASC group and 3 in the NT group identified as Hispanic/Latino. The majority of participants in both groups identified as non-Hispanic/Latino, with 67 in the ASC group and 101 in the NT group. Ethnicity data was unavailable for 3 participants in the ASC group.

Socioeconomic status, as measured by the Hollingshead Four-Factor Index, yielded a mean family SES score of 50.7 (SD = 8.0) for the ASC group and 53.5 (SD = 7.9) for the NT group.
}
\begin{table}[htbp]
\centering
\caption{\textbf{Participant Demographics and Family Socioeconomic Status.} ASC = Autism Spectrum Condition; NT = Neurotypical; SES = Socioeconomic Status.}
\label{tab:participant_demographics}
\begin{tabular}{lcc}
\toprule
 & \textbf{ASC} & \textbf{NT} \\
 & \textbf{Mean (SD)} & \textbf{Mean (SD)} \\
\midrule
Group (n) & 81 & 104 \\
Age (years) & 10.3 (1.7) & 10.2 (1.4) \\
Sex (M:F; n) & 70:11 & 74:30 \\
\midrule
\multicolumn{3}{l}{\textbf{Race (n)}} \\
\hspace{1em}White & 58 & 82 \\
\hspace{1em}African American & 12 & 8 \\
\hspace{1em}Asian & 1 & 6 \\
\hspace{1em}Multiracial & 10 & 8 \\
\midrule
\multicolumn{3}{l}{\textbf{Ethnicity (n)}} \\
\hspace{1em}Hispanic/Latino & 11 & 3 \\
\hspace{1em}Not Hispanic/Latino & 67 & 101 \\
\hspace{1em}Unknown & 3 & 0 \\
\midrule
Hollingshead SES Family & 50.7 (8.0) & 53.5 (7.9) \\
\bottomrule
\end{tabular}
\end{table}

\re{
\subsection{Detailed Movement Types}
\label{sec:movement_types}
Table~\ref{tab:movement_types} provides an overview of the movement types included in Sequence~1 and Sequence~2.

\begin{table}[h]
\caption{Overview of movement types used in Sequence 1 and Sequence 2.}
\label{tab:movement_types}
\centering
\begin{tabular}{ll}
\toprule
\textbf{Sequence 1} & \textbf{Sequence 2} \\
\midrule
1. March & 1. March w/ hand flap \\
2. Roll arms bouncing & 2. Hand Jive \\
3. Heisman to the left & 3. Left hand out \\
4. Vogue Guy & 4. Windmill \\
5. Hands at shoulders & 5. X Hands \\
6. Step half moon & 6. Butterfly arms \\
7. Trombone & 7. T clap \\
8. March down & 8. Star grab \\
9. Rainbow hands & 9. Teapot \\
10. Stop signal & 10. Stop signal \\
11. Chief arms in front of chest & 11. Push pull \\
12. Puppet arms & 12. Star grab \\
13. Rainbow hands & 13. Muscle show to the left \\
14. Hippie to the left & 14. Muscle show to the right \\
15. Hippie to the right & \\
16. Heisman to the right & \\
17. Vogue Guy & \\
18. Starfish & \\
\bottomrule
\end{tabular}
\end{table}

\subsubsection{Sequence 1 -- Movements and Steps}
The following section details the step-level descriptions for the 18 movement types in Sequence~1.

\noindent\textbf{1. March.}
1) Left arm up; 2) Right arm down; 3) Left leg straight; 4) Bend right leg; 5) Left arm trajectory down; 6) Right arm trajectory up; 7) Straighten right leg; 8) Bend left leg; 9) Left arm up; 10) Right arm down; 11) Right leg bent; 12) Straighten left leg; 13) Left arm trajectory down; 14) Right arm trajectory up; 15) Straighten right leg; 16) Bend left leg.

\medskip
\noindent\textbf{2. Roll arms bouncing.}
1) Roll arms over each other away from body; 2) Roll arms over each other away from body; 3) Roll arms over each other away from body; 4) Bounce legs.

\medskip
\noindent\textbf{3. Heisman to the left.}
1) Right hand by ear, facing face; 2) Left hand straight out to left; 3) Left wrist at around shoulder height; 4) Left leg step apart.

\medskip
\noindent\textbf{4. Vogue Guy.}
1) Right arm trajectory straight out to the front; 2) Right wrist at or above hip height; 3) Left hand bent back over shoulder; 4) Left elbow in front of body; 5) Left arm trajectory straight out to the front; 6) Left wrist at or above hip height; 7) Right hand bent back over shoulder; 8) Right elbow in front of body; 9) Right arm trajectory straight out to the front; 10) Right wrist at or above hip height; 11) Left hand bent back over shoulder; 12) Left elbow in front of body.

\medskip
\noindent\textbf{5. Hands at shoulders.}
1) Left hand bent back over shoulder; 2) Left elbow in front of body; 3) Bring right hand over shoulder; 4) Both elbows in front of the body; 5) Step legs together; 6) Both arms trajectory down to straighten along body.

\medskip
\noindent\textbf{6. Step half moon.}
1) Right arm down on the side; 2) Step left leg out to left; 3) Sweep left arm away from the body to the left; 4) Keep left arm in front of body while sweeping; 5) Step right leg in, feet together; 6) Left arm down on the side; 7) Step right leg out to left; 8) Sweep right arm away from the body to the left; 9) Keep right arm in front of body while sweeping; 10) Step left leg in, feet together.

\medskip
\noindent\textbf{7. Trombone.}
1) Bend both elbows; 2) Bring arms close in front of chest; 3) Raise both arms above head, straight; 4) Bend both elbows; 5) Bring arms close in front of chest; 6) Move both arms out in cross position; 7) Bend both elbows; 8) Bring arms close in front of chest.

\medskip
\noindent\textbf{8. March down.}
1) Left arm bent back, fist at hip; 2) Right arm straight down towards knee; 3) Left leg straight; 4) Bend right leg; 5) Right arm bent back, fist at hip; 6) Left arm straight down towards knee; 7) Straighten right leg; 8) Bend left leg.

\medskip
\noindent\textbf{9. Rainbow hands.}
1) Step left leg out; 2) Swing both arms to the left above head; 3) Step right leg next to left leg; 4) Bend right arm from elbow; 5) Right hand in front of neck; 6) Left arm out left, straight; 7) Left wrist at around shoulder height; 8) Step right leg out; 9) Swing both arms to the right above head; 10) Step left leg next to right leg; 11) Bend left arm from elbow; 12) Left hand in front of neck; 13) Right arm out right straight; 14) Right wrist at around shoulder height; 15) Step left leg out; 16) Swing both arms to the left above head; 17) Step right leg next to left leg; 18) Bend right arm from elbow; 19) Right hand in front of neck; 20) Left arm out left straight; 21) Left wrist at around shoulder height; 22) Step right leg out; 23) Swing both arms to the right above head; 24) Step left leg next to right leg; 25) Bend left arm from elbow; 26) Left hand in front of neck; 27) Right arm out right, straight; 28) Right wrist at around shoulder height.

\medskip
\noindent\textbf{10. Stop signal.}
1) Bring both arms straight out in cross position; 2) Lift right arm straight above head; 3) Step right leg to the right; 4) Move left arm in arc to stop position in front of body; 5) Move right arm down next to left arm.

\medskip
\noindent\textbf{11. Chief arms in front of chest.}
1) Bend both arms from elbows stacked in front of chest, left arm over right arm; 2) Lift left arm 90$^\circ$; 3) Lift right arm 90$^\circ$; 4) Step right leg in, legs together.

\medskip
\noindent\textbf{12. Puppet arms.}
1) Both arms bent at elbow, in front of chest; 2) Move right arm down; 3) Move left arm up; 4) Move left arm down; 5) Move right arm up; 6) Move right arm down; 7) Move left arm up; 8) Move left arm down; 9) Move right arm up.

\medskip
\noindent\textbf{13. Rainbow hands (repeat).}
1) Step left leg out; 2) Swing both arms to the left above head; 3) Step right leg next to left leg; 4) Bend right arm from elbow; 5) Right hand in front of neck; 6) Left arm out left, straight; 7) Left wrist at around shoulder height; 8) Step right leg out; 9) Swing both arms to the right above head; 10) Step left leg next to right leg; 11) Bend left arm from elbow; 12) Left hand in front of neck; 13) Right arm out right straight; 14) Right wrist at around shoulder height; 15) Step left leg out; 16) Swing both arms to the left above head; 17) Step right leg next to left leg; 18) Bend right arm from elbow; 19) Right hand in front of neck; 20) Left arm out left straight; 21) Left wrist at around shoulder height; 22) Step right leg out; 23) Swing both arms to the right above head; 24) Step left leg next to right leg; 25) Bend left arm from elbow; 26) Left hand in front of neck; 27) Right arm out right, straight; 28) Right wrist at around shoulder height.

\medskip
\noindent\textbf{14. Hippie to the left.}
1) Both arms down on the sides; 2) Step left leg to left; 3) Turn whole body to the left; 4) Bend both knees; 5) Raise left arm bent at elbow; 6) Right arm straight towards knee; 7) Lean torso forwards; 8) Raise right arm bent at elbow; 9) Left arm straight towards knee; 10) Lean torso backwards; 11) Raise left arm bent at elbow; 12) Right arm straight towards knee; 13) Lean torso forwards; 14) Turn whole body to center; 15) Step legs together.

\medskip
\noindent\textbf{15. Hippie to the right.}
1) Step right leg to right; 2) Turn whole body to the right; 3) Bend both knees; 4) Raise right arm bent at elbow; 5) Left arm straight towards knee; 6) Lean torso forwards; 7) Raise left arm bent at elbow; 8) Right arm straight towards knee; 9) Lean torso backwards; 10) Raise right arm bent at elbow; 11) Left arm straight towards knee; 12) Lean torso forwards; 13) Turn whole body to center; 14) Step legs together.

\medskip
\noindent\textbf{16. Heisman to the right.}
1) Right hand by ear, facing face; 2) Left hand straight out to left; 3) Left wrist at around shoulder height; 4) Left leg step apart.

\medskip
\noindent\textbf{17. Vogue Guy (repeat).}
1) Right arm trajectory straight out to the front; 2) Right wrist at or above hip height; 3) Left hand bent back over shoulder; 4) Left elbow in front of body; 5) Left arm trajectory straight out to the front; 6) Left wrist at or above hip height; 7) Right hand bent back over shoulder; 8) Right elbow in front of body; 9) Right arm trajectory straight out to the front; 10) Right wrist at or above hip height; 11) Left hand bent back over shoulder; 12) Left elbow in front of body.

\medskip
\noindent\textbf{18. Starfish.}
1) Both arms together in front of chest; 2) Kick right leg straight out; 3) Move left arm above head, straight out; 4) Move right arm to right of body, straight out; 5) Right wrist at or above hip height; 6) Return right leg to center; 7) Both arms together in front of chest; 8) Kick left leg straight out; 9) Move right arm above head, straight out; 10) Move left arm to left of body, straight out; 11) Left wrist at or above hip height; 12) Return left leg to center; 13) Both arms together in front of chest; 14) Bounce the knees.

\subsubsection{Sequence 2 -- Movements and Steps}
The following section details the step-level descriptions for the 14 movement types in Sequence~2.

\medskip
\noindent\textbf{1. March w/ hand flap.}
1) Raise both arms up to the left; 2) Arms above head; 3) Bend right leg; 4) Left leg straight; 
5) Lower both arms down to the right; 6) Arms by right hip; 7) Straighten right leg; 8) Bend left leg;
9) Raise both arms up to the left; 10) Arms above head; 11) Bend right leg; 12) Left leg straight;
13) Lower both arms down to the right; 14) Arms by right hip; 15) Straighten right leg; 16) Bend left leg.

\medskip
\noindent\textbf{2. Hand Jive.}
1) Raise both arms out to side; 2) Bend both arms at elbows; 3) Bring right hand over left hand in toward body; 
4) Bring both arms out to side; 5) Bring right hand over left hand in toward body; 6) Bring both arms out to side; 
7) Bring left hand over right hand in toward body; 8) Bring both arms out to side; 
9) Bring left hand over right hand in toward body; 10) Bring both arms out to side.

\medskip
\noindent\textbf{3. Left hand out.}
1) Keep right arm across body; 2) Step left leg to the left; 3) Bring left arm out to the left; 4) Lift right arm straight above.

\medskip
\noindent\textbf{4. Windmill.}
1) Left arm out to the left; 2) Right arm straight above head; 3) Left arm straight above head; 4) Right arm out to the right; 
5) Right arm straight above head; 6) Left arm out to the left; 7) Left arm straight above head; 
8) Right arm out to the right; 9) Legs open shoulder width.

\medskip
\noindent\textbf{5. X Hands.}
1) Step left leg in; 2) Cross arms in front of body in “X” position, right over left arm.

\medskip
\noindent\textbf{6. Butterfly arms.}
1) Step left leg to the left; 2) Bring both arms out to goal post; 3) Shift entire body to the left; 
4) Step feet together; 5) Bring both arms in front of body; 6) Hands at chin; 7) Shift body facing front; 
8) Step right leg to the right; 9) Bring both arms out to goal post; 10) Shift entire body to the right; 
11) Step feet together; 12) Bring both arms in front of body; 13) Hands at chin; 14) Shift body facing front.

\medskip
\noindent\textbf{7. T clap.}
1) Swirl left arm to the left; 2) Left arm straight out to the left; 3) Left wrist at shoulder height; 
4) Swirl right arm to the right; 5) Right arm straight out to the right; 6) Right wrist at shoulder height; 
7) Bring both arms straight above head; 8) Grasp hands together above head; 9) Bring clasped hands down to chest.

\medskip
\noindent\textbf{8. Star grab.}
1) Step left leg out to the left; 2) Raise both arms straight up to the left; 3) Step right leg to the left leg; 
4) Lower both arms straight down to the right; 5) Step right leg out to the right; 6) Raise both arms straight up to the right; 
7) Step left leg to the right leg; 8) Lower both arms straight down to the left; 9) Repeat sequence mirrored.

\medskip
\noindent\textbf{9. Teapot.}
1) Bring right hand to left hip; 2) Bring left hand to left shoulder; 3) Left elbow out to the left; 
4) Step right leg out to the right; 5) Sweep right arm out to the right, keeping it in front of body; 
6) Right arm straight out to the right at shoulder height; 7) Bring left hand to right hip; 
8) Bring right hand to right shoulder; 9) Right elbow out to the right; 
10) Sweep left arm out to the left, keeping it in front of body; 
11) Left arm straight out to the left at shoulder height; 12) Step right leg in to left leg.

\medskip
\noindent\textbf{10. Stop signal.}
1) Left arm straight out to the left at shoulder height; 2) Lift right arm straight above head; 
3) Lower right arm straight out to the front; 4) Move left arm in arc to stop position in front of body; 
5) Legs together.

\medskip
\noindent\textbf{11. Push pull.}
1) Bring left hand to left shoulder, elbow in front; 2) Right arm straight out to the front at shoulder height; 
3) Bring right hand to right shoulder, elbow in front; 4) Left arm straight out to the front at shoulder height; 
5) Repeat alternating arms; 6) Legs together.

\medskip
\noindent\textbf{12. Star grab (repeat).}
Same as Movement 8 (mirrored repetitions).

\medskip
\noindent\textbf{13. Muscle show to the left.}
1) Left arm in up-L position, elbow out to the left; 2) Right arm in down-L position, elbow out to the right; 
3) Step left leg to the left; 4) Alternate arms up/down L-positions with elbows out; 
5) Step left leg in to right leg.

\medskip
\noindent\textbf{14. Muscle show to the right.}
1) Step right leg out to the right; 2) Left arm in up-L position, elbow out to the left; 
3) Right arm in down-L position, elbow out to the right; 4) Alternate arms up/down L-positions with elbows out; 
5) Step right leg in to left leg.

}

\begin{figure}
\centering
\begin{subfigure}{0.92\linewidth}
  
  \includegraphics[width=\linewidth]{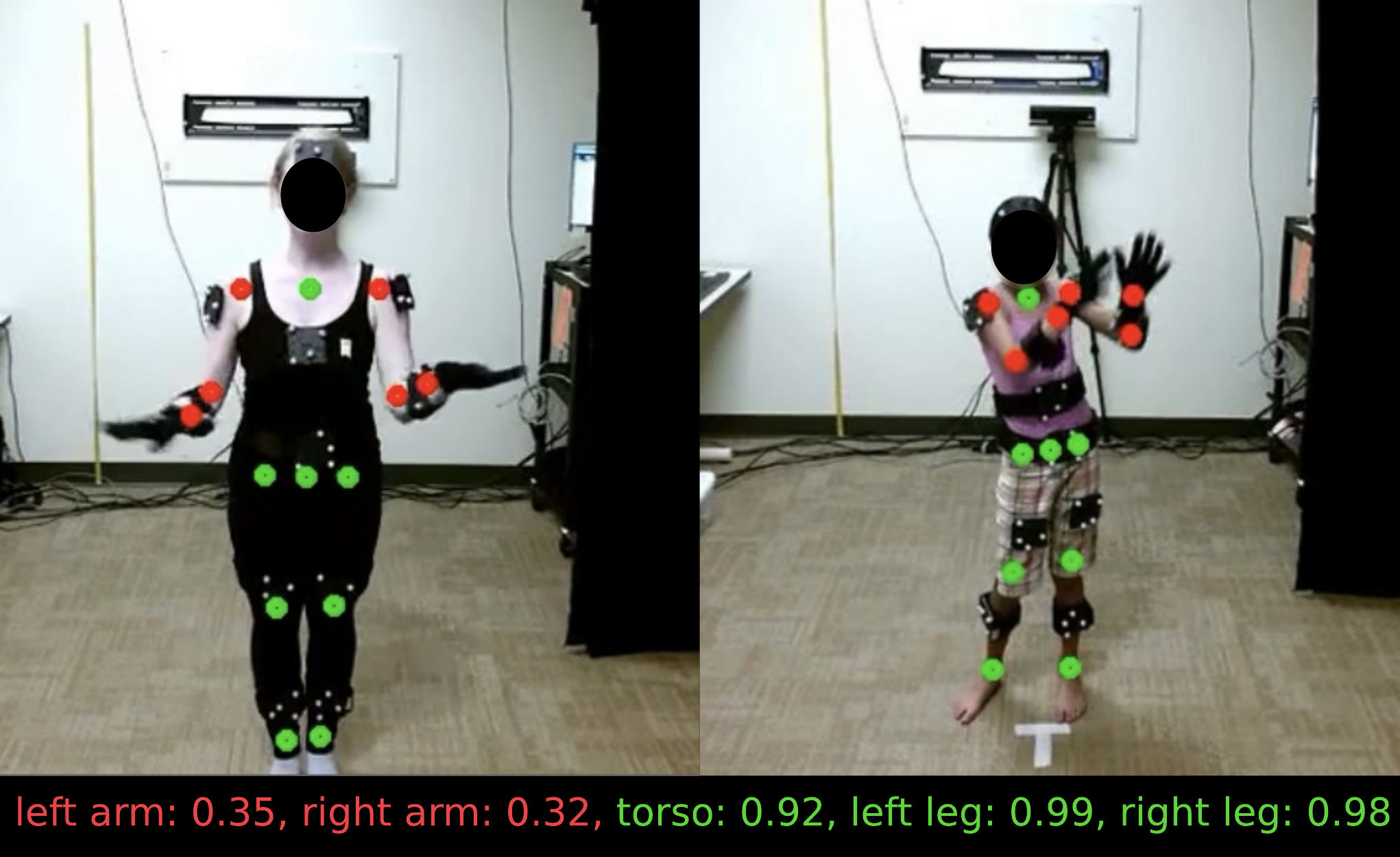}
  \caption{}
\end{subfigure}%
\newline
\begin{subfigure}{0.92\linewidth}
  
  \includegraphics[width=\linewidth]{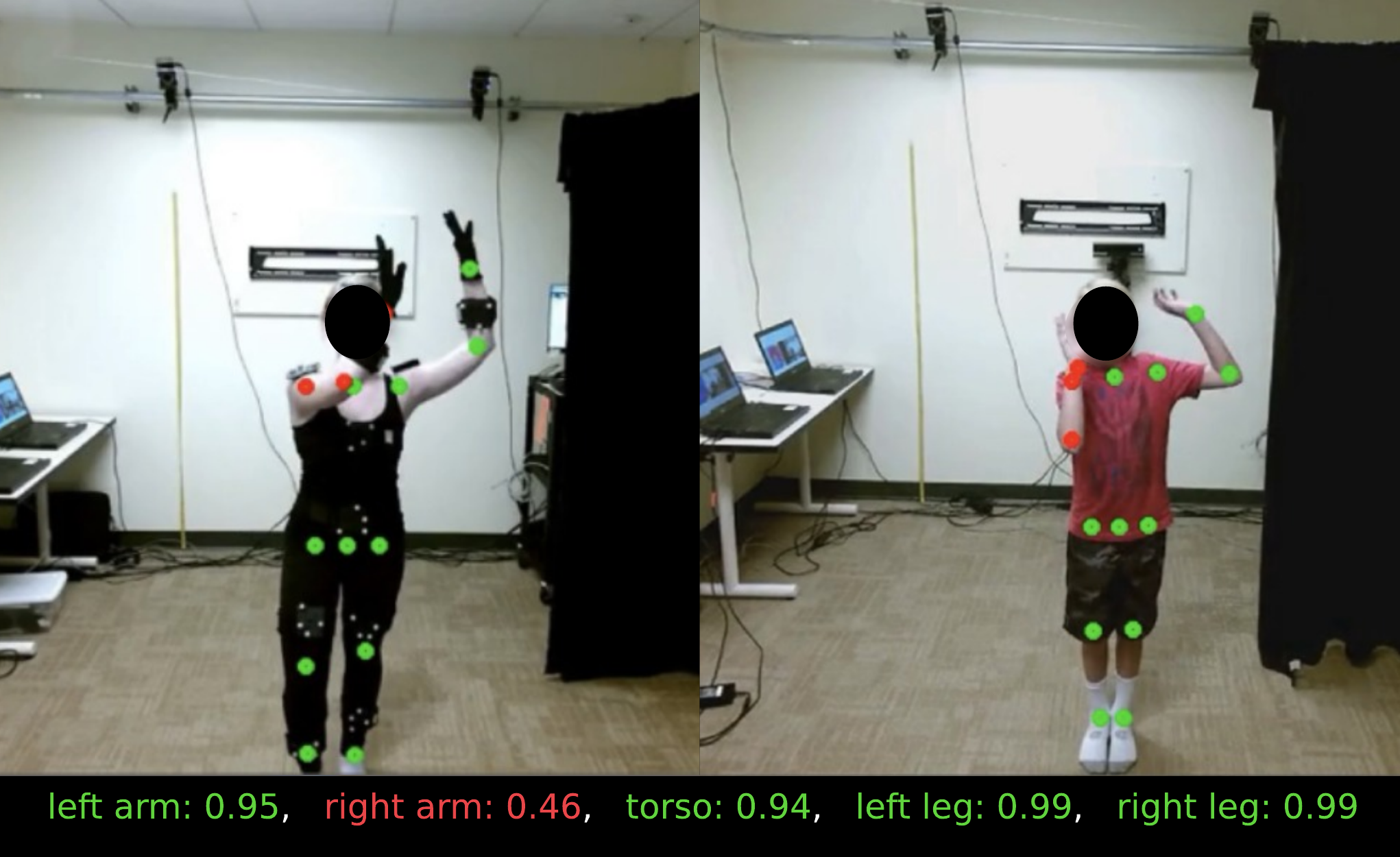}
  \caption{}
\end{subfigure}
\caption{\textbf{Visualization of localized motion imitation scores for body segments, comparing the actor (left) and children's imitation (right).}
(a) Top: The scores indicate low similarity for arms (left: 0.35, right: 0.32), with high alignment for the torso (0.92), left leg (0.99), and right leg (0.98).
(b) Bottom: Higher similarity for the left arm (0.95) but lower for the right arm (0.46). Torso (0.94), left leg (0.99), and right leg (0.99) maintain high alignment. Red highlights low alignment below a threshold, while green indicates high alignment.}
\label{fig:localized_scores}
\end{figure}

\section{Supplementary Experiments}
\label{sec:supplementary_experiments}
\subsection{Interpretable Scores}
\label{sec:interpretable_scores}
As discussed in Section II-B,
to enhance the interpretability of motor imitation assessments, \method\ localizes the analysis to specific body segments, such as the left arm, right arm, torso, left leg, and right leg. By isolating the imitation performance for each segment, \method\ provides detailed insights into which body regions contribute to observed differences in motor imitation. This localized assessment enhances the interpretability of imitation performance, potentially facilitating more targeted interventions.
Figure~\ref{fig:localized_scores} illustrates a snapshot of examples of localized motion imitation scores for two different motions.

\begin{figure}
\centering
\begin{subfigure}{0.99\linewidth}
  
  \includegraphics[width=\linewidth]{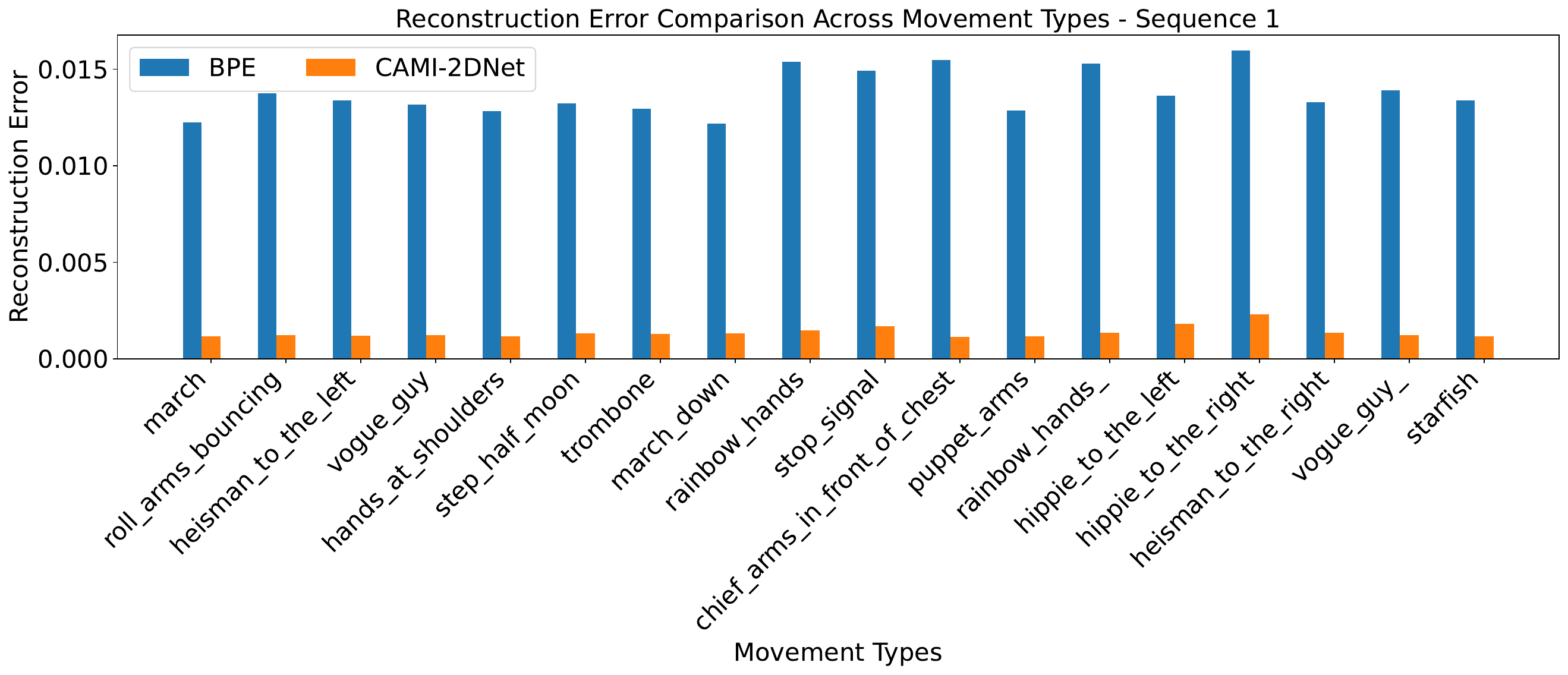}
  \caption{}
\end{subfigure}%
\newline
\begin{subfigure}{0.99\linewidth}
  
  \includegraphics[width=\linewidth]{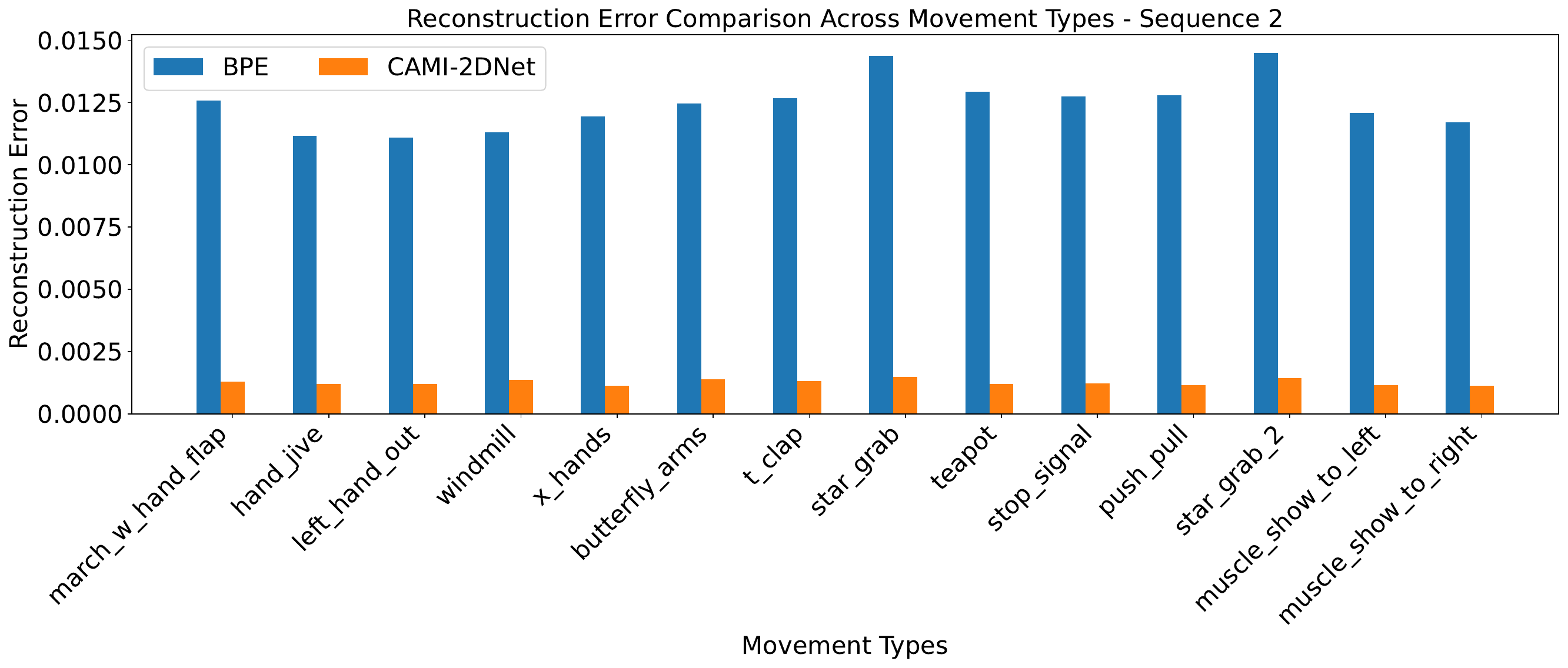}
  \caption{}
\end{subfigure}
\caption{\textbf{Reconstruction error comparison across various movement types for Sequence 1 (a) and Sequence 2 (b).} Lower values indicate better reconstruction quality. Errors are computed in a pixel space normalized to the [0,1] range. \method\ consistently outperforms BPE~\cite{Park2021ABP}, demonstrating greater accuracy and robustness in reconstructing diverse motion patterns in CAMI-185.}
\label{fig:reconstruction_errprs}
\end{figure}

\revision{
\subsection{Evaluation of Reconstruction Quality}
\label{sec:reconstruction_quality}

To comprehensively evaluate the reconstruction quality of motion sequences by our encoder-decoder architecture, we conducted experimental assessments across various movement types in Sequences 1 and 2. Figure~\ref{fig:reconstruction_errprs} illustrates the reconstruction errors for different movement types, comparing the performance of \method\ against a baseline model (BPE~\cite{Park2021ABP}).

\subsection{Evaluation of Different Pose Estimation Networks}
\label{sec:pose_eval}

To assess the robustness of our method regarding the choice of pose estimation network, we conducted inference experiments using three models: EViTPose, HRNet, and HRFormer on the CAMI-47 dataset. As shown in Table~\ref{tab:pose_results}, our method achieves slightly better performance when using the EViTPose network, which it was trained on, compared to HRNet and HRFormer. However, the performance across all three pose estimation methods remains comparable, demonstrating that our approach is robust and independent of the specific pose estimation network employed.

\begin{table}[t]
\centering
\caption{Performance of \method\ with different pose estimation methods on CAMI-47.}
\label{tab:pose_results}
\begin{tabular}{lcccc}
\toprule
\textbf{Pose Estimation Method} & \multicolumn{2}{c}{\textbf{Sequence 1}} & \multicolumn{2}{c}{\textbf{Sequence 2}} \\
\cmidrule(lr){2-3} \cmidrule(lr){4-5}
 & Trial 1 & Trial 2 & Trial 1 & Trial 2 \\
\midrule
EViTPose~\cite{Kinfu2023EfficientVT} & 88.5 & 88.7 & 82.5 & 86.3 \\
HRNet~\cite{Sun2019DeepHR}    & 85.4 & 87.7 & 81.2 & 84.0 \\
HRFormer~\cite{Yuan2021HRFormerHT} & 86.8 & 88.1 & 81.2 & 84.8 \\
\bottomrule
\end{tabular}
\end{table}

\subsection{Evaluation on NTU RGB+D 120 Benchmark}
\label{sec:ntu_rgbd}
To supplement our main experiments, we evaluated the motion similarity modeling capabilities of our approach on the NTU RGB+D 120 benchmark~\cite{Liu2019NTUR1}, using a total of 20,093 motion pairs annotated with human similarity ratings~\cite{Park2021ABP}. Spearman's rank correlation was employed to assess the alignment between the predicted similarity scores and human judgments.

We compared our model against four baseline methods. The first is a simple heuristic that computes Euclidean distances between joints, with Dynamic Time Warping (DTW) used for temporal alignment (Euclidean + DTW). The other three baselines--Coskun~\etal~\cite{Coskun2018HumanMA}, Aberman~\etal~~\cite{Aberman2019LearningCM}, and Park~\etal~\cite{Park2021ABP}--are recent metric learning-based models trained on the SARA dataset, all of which encode entire motion sequences into a single embedding for similarity assessment. As shown in Table~\ref{tab:ntu_rgbd_results}, this evaluation highlights the generalization ability of our model and its competitive performance compared to these state-of-the-art approaches, demonstrating its broader applicability to motion imitation tasks beyond those specific to autism diagnosis.

\begin{table}[t]
\centering
\caption{Spearman correlation between model-predicted motion imitation scores and human annotations on the NTU RGB+D 120 dataset (joints are estimated with MultiPoseNet~\cite{Kocabas2018MultiPoseNetFM} as in~\cite{Park2021ABP}).}
\begin{tabular}{l c}
\toprule
\textbf{Method} & \textbf{Spearman Correlation} \\
\midrule
Euclidean + DTW & 0.2634 \\
Coskun~\etal~\cite{Coskun2018HumanMA} & 0.2845 \\
Aberman~\etal~\cite{Aberman2019LearningCM} & 0.3545 \\
BPE~\cite{Park2021ABP} & 0.5509 \\
\method\ (Ours) & \textbf{0.5690} \\
\bottomrule
\end{tabular}
\label{tab:ntu_rgbd_results}
\end{table}

}

\ree{
\subsection{Ablation on Training Loss Components}

To evaluate the contribution of different components in the training objective, we conducted an ablation study comparing three configurations of the CAMI-2DNet loss function: (i) the full loss used in the main method, combining synthetic and real data losses ($ \mathcal{L}_{\text{total-syn}} + \mathcal{L}_{\text{total-real}}$); (ii) a variant trained only with the synthetic loss ($ \mathcal{L}_{\text{total-syn}}$); and (iii) a variant trained with the full loss but excluding the nuanced motion term from the real loss ($ \mathcal{L}_{\text{total-syn}} + (  \mathcal{L}_{\text{total-real}} -  \mathcal{L}_{\text{total-nuanced}} ) $).

As shown in Table~\ref{tab:loss_ablation}, the full training objective achieves the highest ROC-AUC across both sequences and trials, demonstrating the benefit of training with both synthetic and real motion supervision. Removing the real loss leads to a substantial performance drop, highlighting the importance of domain adaptation to real motor behaviors. When the nuanced motion loss is omitted from the real loss, the performance improves over the synthetic-only variant but remains below the full configuration, indicating that modeling fine-grained motion variations is critical for capturing the richness of human imitation performance.

\begin{table}[ht]
\centering
\caption{Ablation study on training loss components for CAMI-47. The full model combines synthetic and real losses, while the other variants remove or modify specific components.}
\label{tab:loss_ablation}
\begin{tabular}{@{}lcccc@{}}
\toprule
Configuration & \multicolumn{2}{c}{Sequence 1} & \multicolumn{2}{c}{Sequence 2} \\
\cmidrule(lr){2-3}\cmidrule(lr){4-5}
 & Trial 1 & Trial 2 & Trial 1 & Trial 2 \\
\midrule
Full loss (main method) & 88.5 & 88.7 & 82.5 & 86.3 \\
Full without nuanced motion loss & 84.2 & 85.2 & 80.8 & 88.0 \\
Synthetic loss only & 81.2 & 78.3 & 73.7 & 84.0 \\
\bottomrule
\end{tabular}
\end{table}

Overall, these findings confirm that both real-world motion supervision and the nuanced motion term play a crucial role in achieving stable generalization and sensitivity to subtle imitation differences across participants.

}

\ree{
\subsection{Evaluation of Different Body Parts}
To further understand how the inclusion or exclusion of specific body parts affects performance, we evaluated CAMI-2DNet on the CAMI-47 dataset using three configurations: (i) the whole body, (ii) the upper body including the torso, and (iii) the upper body excluding the torso. This analysis helps determine whether full-body motion cues contribute significantly beyond the upper-body dynamics typically emphasized in imitation tasks.
As summarized in Table~\ref{tab:body_part_abl}, the model achieves the highest ROC-AUC when using the whole body, with scores ranging from 82.5 to 88.7 across all sequences and trials. Performance gradually declines as the lower body and torso information are removed, dropping to 81.2-85.3 when only the upper body with torso is used, and further to 77.8-80.8 when the torso is excluded. These results indicate that full-body kinematics provide complementary cues that enhance the model’s ability to capture the nuances of motor imitation, even in tasks that primarily involve upper-body actions.

\begin{table}[ht]
\centering
\caption{ROC-AUC by body region for CAMI-47 across the two sequences and two trials.}
\label{tab:body_part_abl}
\begin{tabular}{@{}l@{\;\;\;}l@{\;\;\;}c@{\;}c@{\;}c@{\;}c@{}}
\toprule
Method & Body region & \multicolumn{2}{c}{Sequence 1} & \multicolumn{2}{c}{Sequence 2} \\
\cmidrule(lr){3-4}\cmidrule(lr){5-6}
 &  & Trial 1 & Trial 2 & Trial 1 & Trial 2 \\
\midrule
CAMI-2DNet & Whole body                    & 88.5 & 88.7 & 82.5 & 86.3 \\
CAMI-2DNet & Upper body (with torso)       & 81.2 & 82.4 & 81.3 & 85.3 \\
CAMI-2DNet & Upper body (without torso)    & 77.8 & 80.8 & 77.1 & 82.5 \\
\bottomrule
\end{tabular}
\end{table}
}

\end{document}